%% file: main.tex
\documentclass[acmsmall,screen,nonacm]{acmart}

\newif\ifarxiv
\arxivtrue

\ifarxiv
  \newcommand{\appcref}[1]{\cref{#1}}
  \newcommand{\appfootnote}{}
\else
  \newcommand{\appcref}[1]{the Appendix}
  \newcommand{\appfootnote}{$^,$\footnote{A full version of our paper with appendices can be found at \url{https://arxiv.org/abs/2504.09246}.}}
\fi

\input{headers/math_commands.tex}

\input{headers/header-lqa}

\input{headers/header}
\input{headers/header-lqa-tail}
\PassOptionsToPackage{numbers, compress, sort, square}{natbib}

\AtBeginDocument{%
  }

\setcopyright{cc}
\setcctype{by-sa}
\acmDOI{10.1145/3729274}
\acmYear{2025}
\acmJournal{PACMPL}
\acmVolume{9}
\acmNumber{PLDI}
\acmArticle{171}
\acmMonth{6}
\received{2024-11-15}
\received[accepted]{2025-03-06}

\begin{document}

\title{Type-Constrained Code Generation with Language Models}

\author{Niels Mündler}
\authornote{Both authors co-lead this project.}
\email{niels.muendler@inf.ethz.ch}
\affiliation{%
  \institution{ETH Zurich}
  \country{Switzerland}
}
\author{Jingxuan He}
\authornotemark[1]
\email{jingxuan.he@berkeley.edu}
\affiliation{%
  \institution{UC Berkeley}
  \country{USA}
}
\author{Hao Wang}
\email{hwang628@berkeley.edu}
\affiliation{%
  \institution{UC Berkeley}
  \country{USA}
}
\author{Koushik Sen}
\email{ksen@berkeley.edu}
\affiliation{%
  \institution{UC Berkeley}
  \country{USA}
}
\author{Dawn Song}
\email{dawnsong@berkeley.edu}
\affiliation{%
  \institution{UC Berkeley}
  \country{USA}
}
\author{Martin Vechev}
\email{martin.vechev@inf.ethz.ch}
\affiliation{%
  \institution{ETH Zurich}
  \country{Switzerland}
}

\renewcommand{\shortauthors}{Niels Mündler*, Jingxuan He*, Hao Wang, Koushik Sen, Dawn Song, Martin Vechev}

\input{sections/abstract.tex}





\maketitle

\input{sections/introduction.tex}

\input{sections/main\_idea.tex}
\input{sections/methods.tex}

\input{sections/typescript.tex}

\input{sections/experiments.tex}
\input{sections/casestudy.tex}
\input{sections/discussion.tex}

\input{sections/related.tex}
\input{sections/conclusion.tex}

\clearpage
\section*{Artifact Availability}
The artifact for this paper, including source code, datasets, and reproductions scripts, is
available on GitHub (\url{https://github.com/eth-sri/type-constrained-code-generation}) and Zenodo \citep{nielsmundler202515232099}.

\section*{Acknowledgements}
We would like to thank the anonymous reviewers for their in-depth and constructive feedback, and the artifact reviewers for their feedback on our artifact accessibility.

\bibliographystyle{ACM-Reference-Format}
\bibliography{references}

\ifarxiv
\input{sections/appendix}
\fi

\end{document}

%% file: headers/math_commands.tex
\usepackage{amsmath,amsfonts,bm}

\def\eqref#1{equation~\ref{#1}}

\def\1{\bm{1}}

\def\vp{{\bm{v}}}

\DeclareMathAlphabet{\mathsfit}{\encodingdefault}{\sfdefault}{m}{sl}
\SetMathAlphabet{\mathsfit}{bold}{\encodingdefault}{\sfdefault}{bx}{n}



%% file: headers/header-lqa.tex
\usepackage[utf8]{inputenc} 
\usepackage{lipsum} 

\usepackage[T1]{fontenc}

\usepackage{etoolbox}
\newbool{includeappendix}
\setbool{includeappendix}{true}

\input{headers/lqa/overfull}



\input{headers/lqa/subfigures}

\input{headers/lqa/colors}

\input{headers/lqa/listing}

%% file: headers/lqa/overfull.tex
\ifdefined\isoverfull
\else
\fi

%% file: headers/lqa/subfigures.tex
\usepackage{subcaption}


%% file: headers/lqa/colors.tex
\usepackage[x11names]{xcolor}

\definecolor{my-full-blue}{HTML}{1F77B4}

\definecolor{my-full-orange}{HTML}{FF7F0E}
\definecolor{my-extra-orange}{HTML}{7C4514}

\definecolor{my-full-green}{HTML}{2CA02C}

\definecolor{my-full-red}{HTML}{d62728}

\definecolor{my-full-purple}{HTML}{9467bd}

\colorlet{my-blue}{my-full-blue!30}
\colorlet{my-orange}{my-full-orange!30}
\colorlet{my-green}{my-full-green!90}
\colorlet{my-red}{my-full-red!90}
\colorlet{my-purple}{my-full-purple!30}

\definecolor{mygreen}{HTML}{B5E3B5}
\colorlet{myred}{LightPink1}
\colorlet{myblue}{SteelBlue2}
\definecolor{mylightblue}{HTML}{B0DBFF}
\colorlet{myorange}{DarkOrange1}

%% file: headers/lqa/listing.tex
\usepackage{listings}

\usepackage{textcomp}

\usepackage{xcolor}

\usepackage[scaled=0.8]{beramono}

\definecolor{ckeyword}{HTML}{7F0055}
\definecolor{ccomment}{HTML}{3F7F5F}
\definecolor{cstring}{HTML}{2A0099}

\lstdefinestyle{numbers}{
	numbers=left,
	framexleftmargin=20pt,
	numberstyle=\tiny,
	firstnumber=auto,
	numbersep=1em,
	xleftmargin=2em
}

\lstdefinestyle{layout}{
	frame=none,
	captionpos=b,
}

\lstdefinestyle{comment-style}{
	morecomment=[l]//,
	morecomment=[s]{/*}{*/},
	commentstyle={\color{ccomment}\itshape},
}

\lstdefinestyle{string-style}{
	morestring=[b]",%
	morestring=[b]',%
	stringstyle={\color{cstring}},
	showstringspaces=false,%
}

\lstdefinestyle{keyword-style}{
	keywordstyle={\ttfamily\bfseries},
	morekeywords={
		function,
		constructor,
		int,
		bool,
		return,
		returns,
		uint
	},
	morekeywords = [2]{},
	keywordstyle = [2]{\text},
	sensitive=true,
}

\lstdefinestyle{input-encoding}{
	inputencoding=utf8,
	extendedchars=true,
	literate=
	{ℝ}{$\reals$}1%
	{→}{$\rightarrow$}1%
	{α}{$\alpha$}1%
	{β}{$\beta$}1%
	{λ}{$\lambda$}1%
	{θ}{$\theta$}1%
	{ϕ}{$\phi$}1%
}

\lstdefinestyle{escaping}{
	moredelim={**[is][\color{blue}]{\%}{\%}},
	escapechar=|,
	mathescape=true
}

\lstdefinestyle{default-style}{
	basicstyle=\fontencoding{T1}\ttfamily\footnotesize,
	style=numbers,
	style=layout,
	style=comment-style,
	style=string-style,
	style=keyword-style,
	style=input-encoding,
	style=escaping,
	tabsize=2,
	upquote=true
}

\lstdefinelanguage{BASIC}{
	language=C++,
	style=default-style
}[keywords,comments,strings]%

%% file: headers/header.tex
\usepackage[english]{babel}
\usepackage{hyperref}
\usepackage{algorithm}
\usepackage{algorithmicx}
\usepackage[noend]{algpseudocode}
\usepackage{color,soul}
\usepackage{ bbold }
\usepackage{multirow}
\usepackage{multicol}
\usepackage{caption}
\usepackage{tabularx,booktabs,xltabular}
\usepackage{graphicx}
\usepackage{tabu}
\usepackage{array}
\usepackage{siunitx}
\usepackage{subcaption}
\usepackage{fontawesome}
\usepackage{stackengine}
\usepackage{svg}
\usepackage{mathtools}
\usepackage{xspace}
\usepackage{wrapfig}
\usepackage{makecell}
\usepackage{placeins}
\usepackage{float}
\usepackage{pifont}
\usepackage{svg}
\usepackage[most]{tcolorbox}
\usepackage{array}
\usepackage{dsfont}
\usepackage{enumitem}
\usepackage[utf8]{inputenc}
\usepackage{textcomp}

\usepackage{syntax}
\usepackage{semantic}

\newcolumntype{x}[2]{S[table-format=#1.#2,table-auto-round]}

\input{headers/lqa/tikz}
\usepackage{pgfplots}
\usepackage[capitalize,noabbrev]{cleveref}
\AtBeginDocument{
\crefname{appendix}{Appendix}{Appendices}
}

\newtheorem{definition}{Definition}

\usepackage{pifont}

\newcommand{\subts}{\ensuremath{L_B}{}}

\definecolor{pastelgreen}{rgb}{0.47, 0.87, 0.47}
\definecolor{pastelred}{rgb}{1.0, 0.6, 0.6}

\usepackage{setspace}

\newcommand{\revisionPC}[1]{{#1}}
\newcommand{\revisionOC}[1]{{#1}}
\newcommand{\revisionRB}[1]{{#1}}

\newcommand{\ttt}[1]{\text{\texttt{#1}}}

\newcommand{\reachable}{\textsc{reachable}}

\newcommand{\derivable}{\textsc{derivable}}
\newcommand{\der}{\derivable}
\newcommand{\depth}{\textsc{depth}}
\newcommand{\roottypes}{\textsc{root}}
\newcommand{\lookup}{\textsc{lookup}}

\newcommand{\te}{\ensuremath{\Gamma}{}}
\newcommand{\ext}{\rightarrowtail}

\makeatletter
\newcommand*{\textoverline}[1]{$\overline{\hbox{#1}}\m@th$}
\makeatother

\newcommand{\states}{\ensuremath{Q}{}}
\newcommand{\trans}{\ensuremath{\delta}{}}
\newcommand{\initial}{\ensuremath{I}{}}

\newcommand{\final}{\ensuremath{F}{}}
\newcommand{\sstate}{\ensuremath{q}{}}
\newcommand{\mstate}{\ensuremath{\mathbf{q}}{}}
\newcommand{\token}{\ensuremath{t}{}}
\newcommand{\sym}{\ensuremath{c}{}}
\newcommand{\istr}{\ensuremath{s}{}}
\newcommand{\graph}{\ensuremath{G}{}}
\newcommand{\reachf}{\ensuremath{\gamma}{}}
\newcommand{\strcat}{\ensuremath{\circ}{}}
\newcommand{\genop}{\ensuremath{\odot}{}}

\newcommand{\sautomaton}{\ensuremath{A}{}}
\newcommand{\sauto}{\sautomaton}

\lstdefinelanguage{JavaScript}{
morekeywords=[1]{break, continue, delete, else, for, function, if, in,
new, return, this, typeof, var, void, while, with, const, let},
morekeywords=[2]{false, null, true, boolean, number, undefined, string,
Array, Boolean, Date, Math, Number, String, Object},
morekeywords=[3]{eval, parseFloat, escape, unescape},
sensitive,
morecomment=[s]{/*}{*/},
morecomment=[l]//,
morecomment=[s]{/**}{*/}, 
morestring=[b]',
morestring=[b]"
}[keywords, comments, strings]

\usepackage{bm}
\newcommand{\markok}{accept}
\newcommand{\markxx}{reject}
\newcommand{\tinyttt}[1]{\footnotesize{\texttt{#1}}}

\usepackage{scalerel}

\newcommand{\crefldecoding}[1]{\begin{NoHyper}\cref{#1}\end{NoHyper}}
\newcommand{\creflreach}[1]{\begin{NoHyper}\cref{#1}\end{NoHyper}}
\newcommand{\wrongtok}[1]{\colorbox{myred}{\hz{#1}}}
\newcommand{\righttok}[1]{\colorbox{mygreen}{\hz{#1}}}

%% file: headers/lqa/tikz.tex
\usepackage{tikz}

\usetikzlibrary{arrows}
\usetikzlibrary{automata}
\usetikzlibrary{calc}
\usetikzlibrary{backgrounds}
\usetikzlibrary{decorations.markings}
\usetikzlibrary{decorations.pathmorphing}
\usetikzlibrary{decorations.pathreplacing}
\usetikzlibrary{fit}
\usetikzlibrary{patterns}
\usetikzlibrary{positioning}
\usetikzlibrary{shadows}
\usetikzlibrary{shapes}
\usetikzlibrary{shapes.geometric}
\usetikzlibrary{arrows.meta}

\definecolor{blue}{HTML}{347bc6}
\definecolor{green-underline}{HTML}{2de12c}
\definecolor{yellow-underline}{HTML}{ffd700}

\tikzstyle{block} = [
    minimum width=1cm,
    minimum height=0.5cm,
    align=center,
    fill=gray!30,
    rounded corners=5pt,
]

\tikzstyle{arrow} = [
	-{Latex[length=1.4mm, width=1.4mm]},
	draw=blue,
]

\tikzstyle{dashedline} = [
    draw=blue,
    dashed
]

\tikzstyle{line} = [
    draw=blue
]

\definecolor{lightblue}{RGB}{173,216,230} 

%% file: headers/header-lqa-tail.tex
\input{headers/lqa/references}

%% file: headers/lqa/references.tex
\usepackage[capitalize]{cleveref}

\crefformat{section}{\S#2#1#3}

\crefrangeformat{section}{\S#3#1#4\crefrangeconjunction\S#5#2#6}

\crefmultiformat{section}{\S#2#1#3}{\crefpairconjunction\S#2#1#3}{\crefmiddleconjunction\S#2#1#3}{\creflastconjunction\S#2#1#3}

\newcommand{\crefrangeconjunction}{--}

\crefname{listing}{Lst.}{listings}
\crefname{line}{Line}{Lines}
\crefname{appendix}{App.}{App.}

\newcommand{\app}[1]{%
	\ifbool{includeappendix}{\cref{#1}}{the appendix}%
}
\newcommand{\App}[1]{%
	\ifbool{includeappendix}{\cref{#1}}{The appendix}%
}

%% file: sections/abstract.tex
\begin{abstract}
Large language models (LLMs) have achieved notable success in code generation. However, they still frequently produce uncompilable output because their next-token inference procedure does not model formal aspects of code. Although constrained decoding is a promising approach to alleviate this issue, it has only been applied to handle either domain-specific languages or syntactic features of general-purpose programming languages. However, LLMs frequently generate code with typing errors, which are beyond the domain of syntax and generally hard to adequately constrain. To address this challenge, we introduce a type-constrained decoding approach that leverages type systems to guide code generation. For this purpose, we develop novel prefix automata and a search over inhabitable types, forming a sound approach to enforce well-typedness on LLM-generated code. We formalize our approach on a foundational simply-typed language and extend it to TypeScript to demonstrate practicality. Our evaluation on the HumanEval and MBPP datasets shows that our approach reduces compilation errors by more than half and significantly increases functional correctness in code synthesis, translation, and repair tasks across LLMs of various sizes and model families, including state-of-the-art open-weight models with more than 30B parameters. The results demonstrate the generality and effectiveness of our approach in constraining LLM code generation with formal rules of type systems.
\end{abstract}

%% file: sections/introduction.tex
\section{Introduction}
\label{sec:introduction}

Large language models (LLMs) are remarkably successful in diverse fields \citep{gpt4-DBLP:journals/corr/abs-2303-08774,few-shot-learners-DBLP:conf/nips/BrownMRSKDNSSAA20,dubey2024llama3herdmodels} and increasingly used in everyday coding tasks \citep{GitHub,Vella2024}. They show promising capabilities at synthesizing code from natural language descriptions \citep{code-llama-DBLP:journals/corr/abs-2308-12950,lozhkov2024starcoder2stackv2}, translating between programming languages \citep{code-llama-DBLP:journals/corr/abs-2308-12950}, and repairing incorrect programs \citep{muennighoff2024octopackinstructiontuningcode,zhang2024survey}. Despite these achievements, LLM-generated code often contains compilation errors, logic flaws, or security vulnerabilities \citep{lostintranslation-DBLP:conf/icse/PanIKSWMSPSJ24,dou2024whatswrongcodegenerated,pearce2022asleep}. These issues arise because LLMs generate code by iteratively sampling the next token from a vocabulary of tokens -- a probabilistic process that does not provide any formal guarantees.

A promising technique to address this limitation is \emph{constrained decoding}, which enforces the formal rules of programming languages during LLMs' code generation process, rejecting invalid tokens and ensuring only valid tokens are considered as generation candidates. Previous studies have shown that constrained decoding improves adherence to program syntax \citep{dominos-DBLP:journals/corr/abs-2403-06988,ugare2024syncode,synchromesh,melcer2024constraineddecodingfillinthemiddlecode}. However, these improvements are limited, as syntax accounts for only a small part of overall program correctness. For instance, in our evaluation of state-of-the-art open-weight LLMs (\cref{sec:experiments}), syntactic errors make up on average $6\%$ of all compilation errors in generated TypeScript code.

\paragraph{Key Challenge: Generating Well-Typed Code}
Beyond program syntax, type systems detect and reject bugs at compile time \citep{type-systems-MITCHELL1990365,matsakis2014rust} and are therefore enforced in many popular programming languages \citep{understandingTypeScript-DBLP:conf/ecoop/BiermanAT14,donovan2015go,arnold2000java}. We observe that LLMs struggle to generate well-typed code \citep{dou2024whatswrongcodegenerated,tambon2024bugs,compilation-benchmark}, as typing rules significantly complicate the generation of valid code \citep{szabo2024compilationquotientcqmetric}. In our evaluation of LLMs (\cref{sec:experiments}), on average $94\%$ of compilation errors result from failing type checks. This suggests a promising direction: guiding LLMs' code generation process by incorporating the formal rules of type systems.

However, implementing this approach is challenging because type systems can in general not be captured by context-free grammars \citep{type-systems-MITCHELL1990365}, prohibiting the application of prior constrained decoding methods developed for program syntax~\citep{ugare2024syncode,dominos-DBLP:journals/corr/abs-2403-06988}. Furthermore, besides deriving and maintaining a type environment for completed expressions during generation (similar to classic type systems), we need to accurately assess and handle partial expressions. Specifically, for each currently generated partial expression, we must decide whether the partial expression can be completed to match a required type.
Determining this would allow us to constrain the LLM to provably generate well-typed expressions upon termination, but involves solving the challenging problem of type inhabitation \citep{inhabitation,completion-DBLP:conf/pldi/GveroKKP13} in the novel context of LLM-based code generation.

\paragraph{This Work: Type-Constrained Decoding}
In this work, we introduce \emph{type-constrained} decoding\footnote{\revisionPC{Our code implementation is publicly available at \url{https://github.com/eth-sri/type-constrained-code-generation}}.}\appfootnote{}, addressing the challenge of generating well-typed code using LLMs . We develop a sound algorithm to determine if a partial program can be completed into a well-typed program. This algorithm is based on a novel non-deterministic automaton we construct. The automaton incrementally builds abstract syntax trees described by the partial program and annotates them with type-relevant context, e.g., declared identifiers and expression types. It leverages such information to maintain a \emph{prefix property}, ensuring that parsing a program prefix only results in a non-empty set of states when it can be completed into a well-typed program. To guarantee the prefix property, we design a sound type search algorithm that determines whether a partial expression can inhabit a given type. We construct our automaton for a generic, simply-typed Turing-complete calculus \citep{understandingTypeScript-DBLP:conf/ecoop/BiermanAT14}.

To demonstrate its practical effectiveness, we instantiate our approach on a non-trivial subset of TypeScript. \revisionPC{We choose TypeScript for three key reasons: (i) it is currently one of the most actively used languages, e.g., in open-source projects on GitHub \citep{Madnight2024,2022topprogramminggithub}; (ii) as we show, state-of-the-art LLMs fail to reliably generate well-typed TypeScript code; (iii) its core type system is simple enough~\citep{understandingTypeScript-DBLP:conf/ecoop/BiermanAT14} to be suitable for developing the first prototype of our approach.} We perform a comprehensive evaluation on TypeScript versions of the widely-used HumanEval and MBPP benchmarks \citep{multiple-DBLP:journals/tse/CassanoGNNPPYZAFGGJ23,humaneval,austin2021programsynthesislargelanguage-mbpp}, focusing on three common coding tasks: synthesis, translation, and repair. Our experimental results show that type-constrained decoding significantly enhances code generation for LLMs of various sizes (2B-34B parameters). For synthesis and translation, it reduces compilation errors by more than half and increases functional correctness relatively by 3.5\% to 5.5\%. Additionally, it enhances functionally correct repair of non-compiling code relatively by 37\% on average. We further investigate our approach in depth through a runtime analyses and case studies.

We highlight that our type constraining approach is broadly applicable to any language derivable from the core calculus, any code generation task in these languages, and any LLM utilizing next-token generation. In \cref{sec:discussion}, we envision how our approach can benefit other production-ready languages and closed-weight LLMs.

\paragraph{Main Contributions}
Our main contributions can be summarized as follows:
\begin{itemize}[leftmargin=7mm]
\item A prefix automaton and a type search algorithm to enable type constraining for LLM-based code generation, demonstrated on a generic, simply-typed core calculus (\cref{sec:method}).
\item An instantiation and extension of our approach to the popular TypeScript language (\cref{sec:typescript}).
\item An extensive evaluation across various LLMs and coding tasks, showing the significant benefit of our approach in reducing compilation errors and increasing functional correctness (\cref{sec:experiments}).
\end{itemize}

%% file: sections/main_idea.tex
\section{Background and Overview}
\label{sec:background}

In this section, we first provide relevant background on LLM-based code generation and constrained decoding.
Then, we motivate our type constraining approach using an illustrative example and present a high-level overview of its construction.

\revisionOC{
\subsection{Background on LLM-based Code Generation and Constrained Decoding}
}
\label{sec:model-constraining}

\begin{wrapfigure}[19]{R}{0.5\textwidth}
  \begin{minipage}{\linewidth}
      \vspace{-6mm}
  \begin{algorithm}[H]
      \caption{\revisionPC{Vanilla LLM-based code generation (without the \textcolor{blue}{blue} highlights) vs. constrained decoding (with the \textcolor{blue}{blue} highlights)}}
      \label{alg:decoding}
      \begin{algorithmic}[1]
      \Require LLM, prompt $x$, \textcolor{blue}{completion engine $CE_L$ for language $L$}
      \Ensure Program $\istr$ \textcolor{blue}{such that $\istr \in L$}
      \State initialize $\istr$ \label{line:decoding-init}
      \State \textbf{while true do} \label{line:decoding-loop1}
          \State \hspace*{\algorithmicindent} $\vp \coloneq \text{LLM}(x \strcat \istr)$ \label{line:decoding-llm}
          \State \hspace*{\algorithmicindent} \textcolor{blue}{\textbf{while true do}} \label{line:decoding-loop2}
          \State \hspace*{\algorithmicindent} \hspace*{\algorithmicindent} $t \sim \vp$ \label{line:decoding-sample}
          \State \hspace*{\algorithmicindent} \textcolor{blue}{\hspace*{\algorithmicindent} \textbf{if} $CE_L(\istr \strcat t)$ \textbf{then break}} \label{line:decoding-ce1}
          \State \hspace*{\algorithmicindent} \textcolor{blue}{\hspace*{\algorithmicindent} \textbf{elif} $t = EOS$ \textbf{and} $\istr \in L$ \textbf{then break}} \label{line:decoding-ce2}
          \State \hspace*{\algorithmicindent} \textcolor{blue}{\hspace*{\algorithmicindent} \textbf{else} $\vp[t] \coloneq 0$; normalize $\vp$} \label{line:decoding-reset}
          \State \hspace*{\algorithmicindent} \textbf{if} $t = EOS$ \textbf{then break} \label{line:decoding-eos}
          \State \hspace*{\algorithmicindent} $\istr \coloneq \istr \strcat t$ \label{line:decoding-cat}
      \State \textbf{return} $\istr$ \label{line:decoding-return}
      \end{algorithmic}
  \end{algorithm}
  \end{minipage}
\end{wrapfigure}

\paragraph{LLM-based Code Generation}
LLMs generate code incrementally by sampling one token at a time in an iterative manner, as depicted in \cref{alg:decoding} (without the blue highlights). A user prompt $x$ specifies a code generation task for a trained LLM. At \crefldecoding{line:decoding-init}, the output program $\istr$ is initialized to an empty string or a program prefix provided in $x$, e.g., a function signature. \revisionPC{At the beginning of each generation iteration (\crefldecoding{line:decoding-llm}), the LLM takes as input a concatenation $x \strcat \istr$ of the prompt $x$ and the current partial program $\istr$. It then predicts a probability distribution $\vp$ over a fixed, finite set of \emph{tokens}, the \emph{vocabulary}, where each token may be a single Unicode character or a string of multiple characters. All common singleton characters are included in LLMs' vocabulary, ensuring that any standard program can be produced by concatenating tokens~\citep{DBLP:conf/acl/SennrichHB16a}. Next, based on distribution~$\vp$, a token~$t$ is sampled (\crefldecoding{line:decoding-sample}) and appended to the program $\istr$ (\crefldecoding{line:decoding-cat}).} This process is repeated until we encounter the special token $EOS$ which signifies the end of the sequence (\crefldecoding{line:decoding-eos}).

LLMs learn to predict adequate probability distributions from extensive training on natural and programming languages \citep{code-llama-DBLP:journals/corr/abs-2308-12950,qwen2025qwen25technicalreport,few-shot-learners-DBLP:conf/nips/BrownMRSKDNSSAA20}.
These distributions implicitly encode language rules, allowing LLMs to successfully solve code generation tasks~\citep{multiple-DBLP:journals/tse/CassanoGNNPPYZAFGGJ23,code-llama-DBLP:journals/corr/abs-2308-12950,deepseek-coder-DBLP:journals/corr/abs-2401-14196}.
However, LLMs may fail to infer complex rules \citep{dyck,formallimitations,maskedstar}, derive incomplete rules for less common languages \citep{multiple-DBLP:journals/tse/CassanoGNNPPYZAFGGJ23,language-distribution-pmlr-v202-orlanski23a}, and, due to the probabilistic nature of its generation procedure, not consistently follow formal language rules.

\paragraph{Constrained Decoding}
The aforementioned shortcoming of LLMs can be mitigated by employing \emph{constrained decoding}, which analyzes the intermediate model outputs $\istr$ during the generation process and enforces that only valid tokens are incorporated.
Specifically, constrained decoding leverages a \emph{completion engine} $CE_L$, specific to a language $L$. Computing $CE_L(s)$ returns whether partial program $\istr$ can be completed to a well-formed program in $L$, meaning whether there exists a (possibly empty) string $\istr'$ such that $\istr \strcat \istr' \in L$.
Equivalently, $CE_L(\istr)$ determines whether $\istr$ belongs to the \emph{prefix language} $L^p$ of $L$, i.e., whether $s \in L^p$. $L^p$ is formally defined as follows:
\begin{definition}
  \label{def:prefix-lang}
  For a given language $L$, its prefix language is $L^p \coloneq \{\istr \mid \exists \istr': \istr \strcat \istr' \in L \}$.
\end{definition}

As illustrated in blue highlights of \cref{alg:decoding}, constrained decoding differs from vanilla LLM-based code generation by adding an additional sample-and-check loop at \crefldecoding{line:decoding-loop2} around the token sampling process at \crefldecoding{line:decoding-sample}.
A sampled token $t$ is considered further only if $s \strcat t$ can be completed to a well-formed program (\crefldecoding{line:decoding-ce1}) or $t$ is $EOS$ and $s$ is already well-formed in $L$ (\crefldecoding{line:decoding-ce2}).
\revisionPC{Otherwise, the probability of $t$ is set to zero at \crefldecoding{line:decoding-reset}, and the sample-and-check loop repeats.
Note that a token $t$ satisfying either \crefldecoding{line:decoding-ce1} or \crefldecoding{line:decoding-ce2} always exists, because $s$ is in $L^p$ and LLMs' vocabulary contains all common characters.
Therefore, the number of iterations of the loop at \crefldecoding{line:decoding-loop2} is bounded by the fixed LLM vocabulary size.}
\revisionPC{In practice, only few iterations are needed (\cref{sec:experiments-runtime}) and do not require additional LLM inference, ensuring a reasonable runtime overhead compared to vanilla decoding.}

The token-level guarantees extend inductively to guarantee the final program's validity with respect to~$L$.
At \crefldecoding{line:decoding-init}, we start with a valid prefix in $L^p$, i.e., either an empty string or a valid prefix provided in the user prompt.
The check at \crefldecoding{line:decoding-ce1} ensures that all intermediate outputs $\istr$ are prefixes in $L^p$.
\revisionPC{Additionally, \crefldecoding{line:decoding-ce2} and \crefldecoding{line:decoding-eos} ensure that the return statement in \crefldecoding{line:decoding-return} is reached only if $\istr \in L$.}
As an additional benefit, by steering previously ill-formed generations into well-formed ones, constrained decoding also increases the likelihood of generating functionally correct code.

Note that commonly used grammar and type checkers can not be used as a completion engine for constrained decoding. They judge whether a program string $\istr$ is well-formed according to the language $L$, i.e., whether $\istr \in L$. 
When $\istr$ is not a complete program in $L$, but a valid prefix in $L^p$, they return a different output than $CE_L(\istr)$, which is not suitable for use in \cref{alg:decoding}.

\newcommand{\hz}{\vphantom{\parbox[c]{1cm}{\rule{1cm}{1.2mm}}}}

\begin{figure}
  \centering
  \begin{minipage}{0.4\textwidth}
    \vspace{1mm}
    \begin{lstlisting}[language=JavaScript,escapechar=\%,basicstyle=\linespread{1.2}\ttfamily\footnotesize,numbers=none,xleftmargin=1mm]
function is_int(text: string): boolean {
  const num = Number(text);
  return !isNaN(num) && 
    parseInt(num%\colorbox{mylightblue}{\hz\textcolor{black}{<completion>}}%
    \end{lstlisting}
  \end{minipage}
  \hfill
  \begin{minipage}{0.57\textwidth}
    \footnotesize
    \setlength\tabcolsep{2.5pt}
    \begin{tabular}{@{}llcccr@{}}
        & \colorbox{mylightblue}{\hz\tinyttt{<completion>}} & Vanilla & Syntax & Types  & Description \\
        \midrule
        (1) & \colorbox{myred}{\hz\tinyttt{;}}      & \markok & \markxx & \markxx & Syntactically invalid \\
        (2) & \colorbox{myred}{\hz\tinyttt{ber}}           & \markok & \markok & \markxx & Undeclared identifier \\
        (3) & \colorbox{myred}{\hz\tinyttt{()}}        & \markok & \markok & \markxx & Disallowed operator \\
        (4) & \colorbox{myred}{\hz\tinyttt{, 10)}}         & \markok & \markok & \markxx & Invalid argument type \\
        (5) & \colorbox{mygreen}{\hz\tinyttt{.toString()}} & \markok & \markok & \markok & Well-formed option \\
    \end{tabular}
  \end{minipage}
  \vspace{-2mm}
  \caption{Left is a partial TypeScript program derived from instance \#113 of the MBPP benchmark~\citep{austin2021programsynthesislargelanguage-mbpp}, awaiting completion. Right are five completion options: (1)-(4) are invalid and (5) is well-formed. Our type-constrained decoding is the only approach capable of correctly rejecting invalid completions and accepting the valid one.}
  \label{fig:intro-partial-sample}
\end{figure}

\subsection{\revisionOC{Overview of Our Type Constraining Approach}}
\label{sec:motivating-example}

\paragraph{Inadequacy of Syntax-Only Constraining}
To apply the constrained decoding algorithm described in \cref{sec:model-constraining}, one needs to choose a language $L$ and implement the completion engine $CE_L$.
Recent work has explored defining $L$ as the set of syntactically valid programs, thus leveraging the syntactic rules of programming languages for constrained decoding \citep{ugare2024syncode,dominos-DBLP:journals/corr/abs-2403-06988,willard2023efficientguidedgenerationlarge}.
However, the benefits of this approach are limited, because syntax accounts for only a small portion of overall program correctness.
For instance, across our evaluations (\cref{sec:experiments}), only 3.5\% of the functional errors and 6\% of the compilation errors in LLM-generated code are due to syntactic errors. 

We illustrate this limitation using the example in \cref{fig:intro-partial-sample}.
It presents five completion candidates for a partial program: (1)-(4) will lead to compilation errors and only (5) can result in a well-formed program.
Based on syntax, completions that contain line terminations or invalid characters (e.g., \ttt{\$}) could be rejected (1).
However, many other cases, including (2)-(4), do not break syntactic rules but still cause compilation errors.
For instance, candidate (2) results in accessing an undeclared identifier.
In candidate (3), the function call operator will fail at execution time, as \ttt{num} is a number and can not be called.
Candidate (4) passes a value of unexpected format to \ttt{parseInt}, which expects the first argument to be a string.
In this example, (4) is generated by CodeLlama 34B~\citep{code-llama-DBLP:journals/corr/abs-2308-12950}.
Syntax-only constraining accepts this invalid completion, leading to a non-compilable final output.

\paragraph{Our Approach: Leveraging the Type System}
We require stronger constraints to effectively guide the model generation.
Beyond syntax, type systems are commonly utilized in compilers, enforcing semantic rules to detect and reject bugs at compile time~\citep{gao2017type}.
For \cref{fig:intro-partial-sample}, the TypeScript type system would correctly reject code containing erroneous completions (2)-(4).
Therefore, in this work, we propose leveraging type systems in constrained decoding to guide code generation.
Our method accurately detects that only candidate (5) is a valid completion, guiding CodeLlama 34B to adopt this option and complete the program correctly.
As detailed in \cref{sec:experiments}, our experimental results demonstrate that our approach more than halves compiler errors in generated code and consistently increases the proportion of functionally correct programs.

Incorporating typing rules into code generation offers substantial potential but presents a significant challenge.
Previous research has focused primarily on constrained decoding for context-free languages, for which prefixes can be efficiently determined \citep{ugare2024syncode,dominos-DBLP:journals/corr/abs-2403-06988,willard2023efficientguidedgenerationlarge}.
Type systems, however, require language specifications that exceed the capabilities of context-free grammars~\citep{type-systems-MITCHELL1990365}, inhibiting the direct application of prior techniques to type-constrained decoding.
Moreover, determining whether a partially generated expression can be completed to be a well-typed full expression involves not only type checking and inference, as done in traditional compilers, but also addressing type inhabitation \citep{complexity-lambda-calc-inf,inhabitation}.

To address these challenges, we design and implement a practical approach to determine whether a string can be completed to a well-typed program.
We begin by developing a specialized kind of non-deterministic automaton that maintains a \emph{prefix property}, formally defined in \cref{sec:ce-def}.
This property ensures that every reachable state can lead to an accepting state. We leverage this property to build a completion engine for constrained decoding as in \cref{alg:decoding}.
We construct such a completion engine to enforce well-typedness for a simply-typed language \subts{} in \cref{sec:ce-id-lit-typ,sec:ce-expressions,sec:ce-stmts} and extend it to a core subset of TypeScript in \cref{sec:typescript}.
At a high level, the automaton acts as a syntactic parser, additionally maintaining information about initialized variables, enclosing function declarations, and other type-related aspects of the partially parsed syntax tree.
This is possible through dynamically created annotated states that track the additional information.

\begin{wrapfigure}[11]{r}{0.32\textwidth}
  \vspace{-7mm}
  \centering
  \includegraphics[width=0.32\textwidth]{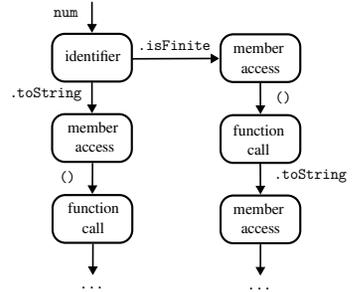}
  \vspace{-7mm}
  \caption{\revisionPC{An example of a prefix automaton.}}
  \label{fig:intro-automaton-a}
\end{wrapfigure}
In \cref{fig:intro-automaton-a}, we provide a concrete example for our prefix automata.
Every state represents the currently parsed syntactic component and additionally tracks the surrounding typing information.
For example, after parsing the partial program in \cref{fig:intro-partial-sample}, the automaton currently parses an expression as the first argument to function \ttt{parseInt}.
Transitions are annotated with further code completions that are deemed admissable based on the syntax and typing information.
In the first state, the automaton has parsed \ttt{num}, inferring from previous declarations that it represents an identifier of type \ttt{number}.
Based on the signature of the \ttt{parseInt} function call, the required type of the completed argument is \ttt{string}.
The automaton now determines the admissable transitions from the identifier state.
State transitions corresponding to completions (1)-(4) from \cref{fig:intro-partial-sample} are disallowed, as they are determined to violate type rules based on the tracked type information.
Further, the automaton needs to determine which operations on the current expression \ttt{num} of type \ttt{number} can be applied to obtain an expression of type \ttt{string}.
To achieve this, we develop a \emph{type reachability search} algorithm, which finds \ttt{string}-typed expressions \ttt{num.toString()} and \ttt{num.isFinite().toString()}. Therefore, it returns that accesses to members \ttt{.toString} and \ttt{.isFinite} are admissible, resulting in the two depicted transitions with the corresponding labels.
In our experiment, CodeLlama 34B chooses to transition along \ttt{.toString()}, the more likely completion based on its training data.
Note that in our actual automaton formalism, as described at the end of \cref{sec:ce-def}, state transitions are on a character level.
\cref{fig:intro-automaton-a} condenses character-level transitions into string-level transitions for presentation purposes.

\begin{wrapfigure}[12]{r}{0.32\textwidth}
  \centering
  \vspace{-4mm}
  \includegraphics[width=.32\textwidth]{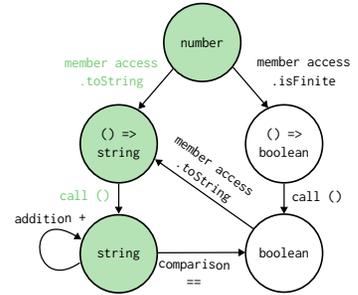}
  \vspace{-7mm}
  \caption{\revisionPC{An example of a partial type search graph.}
  }
  \label{fig:intro-automaton-b} 
\end{wrapfigure}
The type reachability algorithm seeks to identify sequences of operators applied to a given expression such that the resulting expression possesses a required type.
Conceptually, it performs a search over an abstracted type graph, whose nodes are types, and edges represent well-typed operations connecting the input and output types.
An example of such a (partial) graph is shown in \cref{fig:intro-automaton-b}, with a valid path highlighted in green color.
Starting from the derived \ttt{number} type of \ttt{num}, the search first traverses a member access edge to reach the nullary function type \ttt{() => string}.
Then, it traverses an edge representing a function call to reach the goal type \ttt{string}, concluding that the combination of traversed operators \ttt{.toString()} is a well-formed completion for \cref{fig:intro-partial-sample}.
The path for \ttt{num.isFinite().toString()} is analogously valid but omitted in \cref{fig:intro-automaton-b} for brevity.
This type reachability search is invoked every time a partial expression is parsed, in order to determine valid transitions in the prefix automaton.

We implement our approach for a significant subset of TypeScript (\cref{sec:typescript}) and experimentally evaluate it for various LLMs and three important code generation tasks: synthesis, translation, and repair (\cref{sec:experiments}).
The results demonstrate that our approach provides significant benefits in both reducing compilation errors for LLM-generated code and increasing their functional correctness.

%% file: sections/methods.tex
\section{Our Type Constraining Approach}
\label{sec:method}

In this section, we first present a generic, simply-typed language \subts{} (\cref{sec:language}). Then, we present our type constraining approach using \subts{}. Specifically, we introduce our prefix automaton formalism (\cref{sec:ce-def}) and define increasingly complex automata for parsing well-typed fragments of \subts{}, beginning with identifiers, literals, and types (\cref{sec:ce-id-lit-typ}), continuing to expressions, including type search for type-restricted expressions (\cref{sec:ce-expressions}), and concluding with statements (\cref{sec:ce-stmts}).

\subsection{A Simply Typed Language}
\label{sec:language}

We define a simply typed, Turing-complete language, \subts{}.
Its grammar and type system are generic, resembling the principles found in popular statically typed languages, such as TypeScript, Java, and Go.
However, there may be a slight bias towards TypeScript, as our implementation is based on it.

\paragraph{Syntax}
The syntax of \subts{} is shown in \cref{fig:subset-l-detailed}.
The language includes expressions, type-annotated variable and function definitions, and control flows.
Overall, it is based on a core subset of TypeScript~\citep{understandingTypeScript-DBLP:conf/ecoop/BiermanAT14} but can be adapted for other statically typed languages.
Similar to \citet{understandingTypeScript-DBLP:conf/ecoop/BiermanAT14}, we represent Kleene-Star repetitions using an overline, e.g., $\overline{s}$ represents a sequence of statements $s$, and adhere to the TypeScript documentation to annotate parameter types in function signatures with argument names~\citep{tscfunc}.
We make a distinction between base and extension expressions.
The latter applies operators to previous expressions, leading to more complex expressions.
This differentiation is useful later in \cref{sec:ce-expressions} for constructing the prefix automaton for parsing expressions.

\paragraph{Expression Typing Rules}
The typing rules for $\subts$'s expressions are detailed in \cref{fig:inference-rules-expr}.
These rules form a subset of safeFTS, a type-safe portion of TypeScript described by \citet{understandingTypeScript-DBLP:conf/ecoop/BiermanAT14}, allowing us to leverage their soundness results.
The type rules for \subts{} use the standard concept of a \emph{type environment}, denoted as \te, which is a collection of pairs $(x: T)$ of identifiers $x$ and types $T$.
We write $\te \vdash e : T$ if the expression $e$ has type $T$ in the type environment \te.
An expression $e$ is considered valid if its type can be derived by applying the given typing rules.

Literals are evaluated to their respective types ($\textsc{lit}-\{\textsc{num}, \textsc{str}, \textsc{bool}\}$). Identifiers $x$ are evaluated based on the corresponding type in the type environment ($\textsc{ident}$). Anonymous functions are typed according to their annotated parameter types, with the return type determined by the returned expression ($\textsc{anon}$). Grouping preserves the type of the inner expression ($\textsc{group}$).

Binary operators have predefined signatures $S_1 \genop S_2 : T$, such as $\ttt{number}\ \ttt{+}\ \ttt{number} : \ttt{number}$ for addition and $T\ \ttt{=}\ T : T$ for assignments.
These signatures must be satisfied in well-typed expressions ($\textsc{op}$).
Function calls require parameters to match the function signature ($\textsc{call}$). The type of member accesses $e\ttt{.}n$ is determined using an auxiliary function $\lookup(S, n)$, which fetches the type of member $n$ for type $S$. An instantiation of $\lookup$ for TypeScript is provided by \citet{understandingTypeScript-DBLP:conf/ecoop/BiermanAT14}.

\input{figures/grammar/grammar-detailed.tex}

\paragraph{Statements and Type Environments}
The typing rules for statements are presented in \cref{fig:inference-rules-stmt}.
Type environments are modified by statements, in particular variable declarations and function definitions. We use the notation $\te_1 \vdash s \ext \te_2$ to indicate that after executing statement $s$ in type environment $\te_1$, the new environment is $\te_2$.

Variable declarations introduce the identifier with declared type into the type environment, provided the identifier is not already defined ($\textsc{decl}$).
The type environment defines the context to evaluate expressions ($\textsc{expr}$) and return statements ($\textsc{ret}$).
Return statements are only well-typed inside function bodies.
The statements inside statement blocks and if-then-else statements must maintain valid type environments, but do not have an external effect ($\textsc{block}$, $\textsc{ite}$).
This also applies to function definitions; however, the defined function is finally added to the external type environment ($\textsc{fun}$).
Lastly, empty statements do not alter the type environment ($\textsc{nop}$), while statement sequences propagate the type environment along the execution ($\textsc{seq}$).

\input{figures/grammar/inference-expr.tex}
\input{figures/grammar/inference-stmt.tex}
\input{figures/grammar/inference-ret}

\paragraph{Return Types}
The rules for checking return types are presented in \cref{fig:inference-rules-return}. Firstly, return statements must contain expressions matching the function's declared return type.
Secondly, such an expression must be returned on every execution path.
We use the notation $\te \vdash \overline{s} : R$ to indicate the sequence of statements $\overline{s}$ ensures a return value of type $R$.

For variable declarations and expression statements, the return type of the subsequent statements is considered (\textsc{r-decl}, \textsc{r-expr}).
The return type of a return statement directly corresponds to the type of the returned expression (\textsc{r-ret}).
For statement blocks, the return type is decided by either the block itself or the subsequent statements (\textsc{r-block-self}, \textsc{r-block-next}).
In function definitions, the return type is determined by the type of the subsequent statements, similar to expression statements. It is additionally required that the function body returns a type matching the declared return type (\textsc{r-fun}).
For if-then-else statements, both branches must return the same type (\textsc{r-ite-self}), or the return type is determined by the following statements (\textsc{r-ite-next}).

\paragraph{Language Definition}
In summary, a program $s$ is in language \subts{} if both (i) $s$ conforms to the grammar in \cref{fig:subset-l-detailed} and (ii) $s$ is well-typed according to the typing rules in \cref{fig:inference-rules-expr,fig:inference-rules-stmt,fig:inference-rules-return}.

\subsection{Prefix Automaton Definition}
\label{sec:ce-def}

We introduce a general definition of \emph{prefix automata}, beginning with basic automata concepts.
Prefix automata are standard automata that ensure a special \emph{prefix property}\footnote{\revisionRB{Note that the prefix property defined in our work differs from the one discussed in classical texts, e.g., \citep{empty-sensitive-hopcroft-ullmann-DBLP:books/aw/HopcroftU79}}}.
This property enables us to use a prefix automaton to decide whether some string is in the prefix language $L^p$ of a given language $L$.
That is, the prefix automaton can function as a completion engine $CE_L$ to facilitate constrained decoding, as described in \cref{sec:model-constraining}.

\revisionRB{
We consider an automaton $\sautomaton{} \coloneq \langle \Sigma, \states, \trans, \initial, \final\rangle$, a tuple of the five following elements: (i) $\Sigma$ is an alphabet of input symbols; (ii) $\states$ is a set of states; (iii) $\trans: \states \times \Sigma \mapsto \mathcal{P}(\states)$ is a computable transition function that maps a state and an input symbol to a finite set of next states; (iv) $\initial \subseteq \states$ is a finite set of initial states; and (v) $\final \subseteq \states$ is a decidable set of accepting states.
As a convention, we denote a symbol in $\Sigma$ as $\sym$, a string of symbols in $\Sigma^{\ast}$ as $\istr$, the empty string as $\varepsilon$ and an operator for concatenating symbols and strings as $\strcat$.}
The transition function \trans{} maps a given state to all possible subsequent states.
When \trans{} is applied on a set of states $\mstate \subseteq \states$, we take the union of the results as output, i.e., $\trans(\mstate, \sym) \coloneq \bigcup_{\sstate \in \mstate}{\trans(\sstate, \sym)}$. 
The transition function defines a directed graph \graph{} over \states{}, where every state is a node and there is an edge annotated with $\sym$ from $\sstate$ to $\sstate'$ if $\sstate' \in \trans(\sstate, \sym)$.
The language parsed by \sautomaton{} comprises all strings $\istr$ such that traversing $G$ from some initial state in $\initial$ along the edges annotated with $\sym_i$ for $\sym_1\strcat\sym_2\strcat\dots\strcat\sym_n = \istr$, it is possible to reach some accepting state in $\final{}$.
Formally, we define recursively a traversal function \reachf{} for states \mstate{} as $\reachf(\mstate, \varepsilon) \coloneq \mstate$ and $\reachf(\mstate, \istr \strcat \sym) \coloneq \trans(\reachf(\mstate, \istr), \sym)$.
The language accepted by \sautomaton{} is then defined as $L(\sautomaton) \coloneq \{\istr \mid \reachf(\initial, \istr) \cap \final \neq \varnothing\}$.
The traversal function has two intuitive properties concerning reachability that can be shown inductively:

\begin{itemize}[leftmargin=9mm]
\item [(P1)] A path along the graph can be split arbitrarily, i.e., $\reachf(\mstate, \istr \strcat \istr') = \reachf(\reachf(\mstate, \istr), \istr')$. \label{haha}
\item [(P2)] If a state is reached by $\istr \strcat \istr'$, some state is reachable by $\istr$, i.e., $\reachf(\mstate, \istr \strcat \istr') \neq \varnothing \implies \reachf(\mstate, \istr) \neq \varnothing$.
\end{itemize}

An automaton satisfies the \emph{prefix property} or is a \emph{prefix automaton}, if there is a path from every reachable state to some accepting state, or formally:
\begin{definition}
\label{def:prefix}
  For an automaton $\sauto$, the prefix property holds iff \ $\forall \sstate \in \reachf(\initial, \istr): \exists \istr': \reachf(\sstate, \istr') \cap \final \neq \varnothing$. The automaton is a prefix automaton if it satisfies the prefix property.
\end{definition}

Intuitively, for such automata, reaching some state by consuming string \istr{} implies that \istr{} is a prefix to some member of $L(A)$.
We define the reachable language of \sautomaton{}, all inputs that result in some state, as $L_r(\sautomaton) \coloneq \{\istr \mid \reachf(\initial, \istr) \neq \varnothing\}$.
Below, we establish the equivalence of $L_r(\sautomaton)$ and $L(\sautomaton)^p$, the prefix language of $L(\sautomaton)$ as defined in \cref{def:prefix-lang}.

\begin{lemma}
\label{lemma:prefix}
  If \sautomaton{} is a prefix automaton, then $L(\sautomaton)^p = L_r(\sautomaton)$.
\end{lemma}
\begin{proof}
For any $s \in L(\sautomaton)^p$ there exists $s'$ such that $s \strcat s' \in L(\sautomaton)$, by the definition of prefix languages.
By the definition of $L(A)$, this implies $\reachf(\initial, \istr \strcat \istr') \neq \varnothing$.
Then, using (P2), we further derive $\reachf(\initial, \istr) \neq \varnothing$, i.e., $s \in L_r(\sautomaton)$.
Therefore, $L(\sautomaton)^p \subseteq L_r(\sautomaton)$ holds.
The other direction also holds.
We first see that $\istr \in L_r(\sautomaton) \implies \reachf(\initial, \istr) \neq \varnothing$.
Then applying \cref{def:prefix} and (P1), we find $\exists \istr': \reachf(\initial, \istr \strcat \istr') \cap \final \neq \varnothing$, implying $\istr \strcat \istr' \in L(\sautomaton)$ and thus $\istr \in L(\sautomaton)^p$.
\end{proof}

Note that $L(\sautomaton)^p \subseteq L_r(\sautomaton)$ holds generally for automata, since the first half of the proof does not require the prefix property.

\revisionOC{
\paragraph{From Prefix Automata to Completion Engines}
With \cref{lemma:prefix}, given a prefix automaton $A$, we can define a convenient-to-compute completion engine for the underlying language $L(A)$: $CE_{L(\sauto)}(\istr) \coloneq \reachf(\initial, \istr) \neq \varnothing$.
Since our target language is $L$ and not $L(\sauto)$, we now need to determine the relationship between $L(A)$ and $L$.
If we construct $A$ such that it parses a subset of $L$, i.e., $L(A) \subseteq L$, we are guaranteed that all LLM generations constrained by $CE_{L(\sauto)}$ lie in $L$.
Conversely, if $L(A) \supseteq L$, we are guaranteed that every string in $L$ can be expressed under constrained decoding, but not that every generation is valid.
For example, if $A$ permits all syntactically correct programs, it guarantees that all well-typed programs can be generated, but permits ill-typed programs as well.
Therefore, $L(A) \subseteq L$ is required to achieve our goal of enforcing well-typedness on LLM-generated code.
Ideally, $A$ would parse $L$ exactly, i.e., $L(A) = L$, which in our setting additionally guarantees that every well-typed program can be expressed under the constraints of the completion engine.
If this is not achieved, it is important for $A$ to capture a large subset of $L$ to be practically useful.

\paragraph{Building a Prefix Automaton for $L_B$: Warming up}
In the next sections, we will construct a prefix automaton for soundly parsing well-typed programs in $L_B$, by presenting various prefix automata for well-typed fragments of \subts{}.
Our final automaton will cover a significant but incomplete subset of $\subts$.
Incompleteness exists because to ensure that our algorithms terminate, we do not cover high-order types that are less likely to occur in practice.
This is discussed in more detail in \cref{sec:ce-expressions}.
Our evaluation in \cref{sec:experiments} empirically demonstrates that our approach sufficiently covers practical use cases to significantly improve the correctness of LLM-generated code.
}

\revisionPC{We choose $\Sigma$ to be the set of Unicode characters.
This makes our completion engine agnostic to LLM vocabularies.
Even though LLMs' vocabularies differ, their tokens are always a string of single or multiple characters.
When our completion engine for $L_B$ is called during constrained decoding, i.e., at \crefldecoding{line:decoding-ce1} of \cref{alg:decoding}, it processes the sampled token character by character.}

Before proceeding, we briefly introduce several base prefix automata below, with their precise definitions detailed in \appcref{app:ce-basics}.
These automata are later combined, with parts of the transition function being overwritten, to construct more complex automata that capture elements of $L_B$.
\begin{itemize}[leftmargin=7mm]
  \item Union $A_X \cup A_Y$ parses the language $\lbrace \istr \mid \istr \in L(A_X) \cup L(A_Y) \rbrace$. It is a prefix automaton if both $A_X$ and $A_Y$ are prefix automata.
  \item Concatenation $A_{XY}$ parses the language $\lbrace \istr \strcat \istr' \mid \istr \in L(A_X), \istr' \in L(A_Y) \rbrace$. It is a prefix automaton if $A_X$ and $A_Y$ are both prefix automata, and $L(A_Y) \neq \varnothing$.
  \item Kleene-Star $A_{\overline{X}}$ parses the language $\lbrace \overline{\istr} \mid \istr \in L(A_X) \rbrace$. It is a prefix automaton if $A_X$ is a prefix automaton.
  \item Terminal $A_{\ttt{S}}$ parses the language $\{\ttt{S}\}$, where $\ttt{S}$ is a fixed, non-empty string.
  \item Empty $A_{\varnothing}$ parses the empty language $\varnothing$ and is always a prefix automaton.
\end{itemize}

\subsection{Prefix Automata for Identifiers, Literals, and Types}
\label{sec:ce-id-lit-typ}

We now introduce prefix automata for basic syntactic elements of $L_B$: identifiers, literals, and type annotations.
The languages parsed by these automata exactly match their counterparts in $L_B$.

\paragraph{Literals}
The prefix automaton for literals $\sauto_{l} \coloneq \sauto_{\textsc{num}} \cup \sauto_{\textsc{str}} \cup \sauto_{\textsc{bool}}$ accepts number, string, and boolean literals as defined in \cref{fig:subset-l-detailed}. The automata $\sauto_{\textsc{num}}$, $\sauto_{\textsc{str}}$, and $\sauto_{\textsc{bool}}$ are defined by the deterministic finite automaton representation of the corresponding regular expression of the literal. To ensure the prefix property on the finite automata of the regular expression, we prune states from which accepting states can not be reached.

\paragraph{Identifiers}
During parsing, we maintain the current type environment \te{}, as detailed in \cref{sec:ce-stmts}. We define the identifier automaton $A_{x}$ as the union of the terminal automata for identifiers defined in \te{}. In other words, $A_{x} \coloneq \bigcup_{y \in \te}A_{\ttt{y}}$.

\paragraph{Types}
The type automaton $A_{T}$ accepts type annotations as defined in the grammar of \subts~(\cref{fig:subset-l-detailed}). It is defined as $A_{T} \coloneq A_{\textsc{type-lit}} \cup A_{\textsc{type-fun}}$. This includes type literal automaton $A_{\textsc{type-lit}} \coloneq A_{\ttt{string}} \cup A_{\ttt{number}} \cup A_{\ttt{boolean}}$ and function type automaton $A_{\textsc{type-fun}} \coloneq A_{\ttt{(}\overline{p}\ttt{)}\ \ttt{=>}\ T}$. The latter is a concatenation of multiple prefix automata, with the parameter and return types recursing on $A_{T}$.
This recursive definition is valid, since it ensures a finite set of initial states, defines a decidable accepting set, and preserves the prefix property.

\subsection{Prefix Automaton for Expressions}
\label{sec:ce-expressions}

We introduce prefix automata to parse well-typed expressions in $L_B$.
We begin by describing an automaton $A_e$ to parse expressions whose types are unrestricted, e.g., any expression $e$ in an expression statement $e\ttt{;}$.
Then, we present an automaton $A_e \downarrow T$ for expressions whose type is constrained to $T$, e.g., for parameters of function calls.
The type-constrained version accepts a string only if the inhabited type of the represented expression matches $T$.
To preserve the prefix property, we need to ensure that partial expressions can be completed to inhabit the constrained type.
Completions may involve arbitrarily many applications of operators, which may modify the expression type.
We therefore introduce a type search algorithm that soundly determines which types an expression can inhabit, and use it to prune transitions that violate the prefix property. 


\paragraph{Unrestricted Expressions}
\revisionPC{
To handle the recursive syntactic structure of expressions, we differentiate two kinds as shown in \cref{fig:subset-l-detailed}: \emph{base expressions}, including identifiers, literals, grouped expressions, and anonymous functions, and \emph{extension expressions}, which are operator applications (binary operator, member access, or function call) that lead to extending a given expression.
}

The expression automaton $A_e$ is thus defined as the union of base expression automata $A_{x}$, $A_l$, $A_{\ttt{(}e\ttt{)}}$, and $A_{\ttt{(}\overline{p}\ttt{)}\ \ttt{=>}\ e}$, with potential extensions $A_{\genop e}$, $A_{\ttt{.}n}$, and $A_{\ttt{(}\overline{e}\ttt{)}}$.
The individual base and extension automata are constructed by concatenating the respective terminal automata and recursively $A_e$. Additionally, we restrict the type of the recursive $A_e$ if the restriction is required by the type system, e.g., for parsing call parameters with a fixed type. 
We provide additional details on this restriction in \appcref{app:ce-expression-details}.
Since an expression can end after either base or extensions, accepting states of both base and extending automata are accepting states of $A_e$. 
To implement extensions, we start from the base expression automata and recursively adjust $A_e$'s transition function $\trans_e$ by adding outgoing edges from the accepting states of the current automaton to the initial states of the extending automata, or formally:
\[
    \forall X, Y: \trans_{e}(\sstate^X_Y, \sym) \coloneq 
\begin{cases}
    \trans_Y(\sstate^X_Y, \sym) \cup \trans_e(\initial^{X}_{\ttt{(}\overline{e}\ttt{)}}, \sym) \cup \trans_e(\initial^{X}_{\genop e}, \sym) \cup \trans_e(\initial^{X}_{\ttt{.}n}, \sym) & \text{if } \sstate^X_Y \in \final_{Y} \\
    \trans_Y(\sstate^X_Y, \sym) & \text{otherwise,}
\end{cases}
\]
where the labels $X$ and $Y$ for a state $q_Y^X$ represent that a string $X$ has been parsed, and currently the active automaton is $A_Y$, which can be one of the following: $A_{x}$, $A_l$, $A_{\ttt{(}e\ttt{)}}$, $A_{\ttt{(}\overline{p}\ttt{)}\ \ttt{=>}\ e}$, $A_{\genop e}$, $A_{\ttt{.}n}$, and $A_{\ttt{(}\overline{e}\ttt{)}}$.
The superscripts are useful for tracking the currently expressed type, enabling us to determine the validity of extensions and transition to type-restricted expressions based on $L_B$'s typing rules.
For instance, for state $q^{\ttt{42}}$, the addition operator extension $+e$ and function call extension $(\overline{e})$ are syntactically applicable to \ttt{42} of type \ttt{number}.
While the addition operator with type signature \ttt{number}\,\ttt{+}\,\ttt{number}\,:\ttt{number} is allowed, we can not apply a function call on \ttt{number}.
In general, we set $I^X_Y \coloneq \varnothing$ when $Y$ is an invalid extension to $X$.
Moreover, for the extension $+e$ to be valid, $e$ must be of type \ttt{number}.
To this end, we transition to a type-restricted expression automaton by setting $I^{\ttt{42}}_{\ttt{+} e}$ to the set of initial states for $A_{\ttt{+}} \strcat (A_e \downarrow \ttt{number})$. Similar to the recursive type automaton, our definition of $A_e$ ensures a finite set of initial states and a decidable accepting set.




\paragraph{Type-Constrained Expressions}
To implement $A_e \downarrow T$, we must determine whether a partial expression $s$ can be completed to inhabit type $T$.
Completing $s$ without any extension can lead to a possible set of types and repeated extensions can further alter the result type, but we are not guaranteed that the desired type can be reached.
Moreover, extensions can be applied indefinitely, prohibiting an exhaustive search of possible completions.

We therefore develop a two-tiered algorithm, which we describe in the following paragraphs.
This algorithm first identifies the \emph{derivable types} $\derivable(\sstate_s)$ of $s$ based on its current state $\sstate_s$.
$\derivable(\sstate_s)$ refers to the set of inhabitable types for all possible expressions completed from $s$ without extension.
Second, a \emph{type reachability search} $\reachable(\derivable(\sstate_e), T)$ is performed to determine if $T$ can be inhabited by extending from the derivable types of $s$.

We prune automaton transitions when this type search returns a negative result.
To ensure the prefix property, the performed search is \emph{sound}, i.e., it only returns a positive result if $T$ can be expressed by a valid sequence of extensions.
This also aligns with our goal of generating only well-typed programs, ensuring that our expression automata accept a subset of all well-typed expressions of \subts.
To ensure termination, the search is \emph{incomplete}, i.e., there may be a valid sequence of transitions to express $T$ which is not found by the search and we may end up disallowing generation of a well-typed expression.
However, it only avoids traversing types of high complexity that are less likely to occur in practice.
We further empirically ensure that our approach is practically effective (\cref{sec:experiments}).

\begin{wraptable}[13]{r}{0.57\textwidth}
  \vspace{-4mm}
  \small
  \centering
  \renewcommand{\arraystretch}{1.0}
  \caption{Definition of $\der(x)$ for partial expressions introduced in \cref{fig:subset-l-detailed}. $\istr \leq \istr'$ expresses that $\istr$ is a prefix of $\istr'$. $\text{pmatch}(s, T)$ determines whether a prefix $s$ partially matches the regular expression of literals of type $T$.}
  \label{fig:subset-derivable-def}
  \vspace{-3.5mm}
  \begin{tabular}{@{}ll@{}}
    \toprule
    $\istr$ & $\der(\sstate_{\istr})$ \\
    \midrule
    $l$ & $\{T \mid{} \text{pmatch}(l, T), T \in \{\ttt{number}, \ttt{string}, \ttt{boolean}\}\}$ \\
    $x$ & $\{T \mid{} x \leq n, (n: T) \in \te \}$ \\
	  \ttt{(}$\overline{p}$\ttt{)}\ \ttt{=>}\ $e$ & $\{ \ttt{(}\overline{p}\ttt{)}\ \ttt{=>}\ T \mid{} \reachable(\der(\sstate_e), T)\}$ \\
	  \ttt{(}$e$ & $\{T \mid{} \reachable(\der(\sstate_e), T)\}$ \\
    $e\genop$ & $\{T \mid{} \exists S': \te |- e : S \land S \genop S' : T \}$ \\
    $e$\ttt{(} & $\{R \mid{} \te |- e : \ttt{(} \overline{p} \ttt{)}\ \ttt{=>}\ R \}$ \\
    $e$\ttt{.}$a$ & $\{ S \mid{} a \leq n, \te |- e: T, \lookup(T, n) = S\}$ \\
    \bottomrule
  \end{tabular}
\end{wraptable}

\paragraph{Derivable Types}
For the first part of the algorithm, we determine all types inhabitable by the currently parsed expression $s$ without extension, i.e., $\derivable(\sstate_s)$. For example, while parsing partial identifier $x$ in the type environment $\te \coloneq \{(x: \ttt{number}), (xy: \ttt{string})\}$, we have $\der(\sstate_{\ttt{x}}) = \{\ttt{number}, \ttt{string}\}$ and $\der(\sstate_{\ttt{xy}}) = \{\ttt{string}\}$.
For a final state \sstate{} of expression $e$, we define $\der(\sstate) \coloneq T$, where $\te |- e : T$. Different expressions impose different rules on derivability, and we present the detailed rules in \cref{fig:subset-derivable-def}.
Note that for grouped expressions and function literals, we need to enumerate reachable types by recursively contained expressions. 
To avoid explicitly enumerating all reachable types, we integrate the derivability and reachability algorithms.
This optimization is discussed in more detail in \appcref{app:ce-recursion}.

\begin{lemma}
  \label{lemma:derivable}
	For state $\sstate \in \reachf(\initial_e, \istr)$ of partial expression $\istr$, $\der(\sstate)$ returns all $T$ s.t. exists some suffix $\istr'$ with $\te \vdash \istr \strcat \istr' : T$ and $s'$ does not involve an extension (operator, call, or member access).
\end{lemma}
\begin{proof}
  By case distinction on the possible states of partial expressions.
\end{proof}

\paragraph{\revisionOC{Type Reachability}}
To determine which types are inhabitable by extending a base expression $e$ of a given type $T$ (with binary operator, function call, or member access), we analyze sequences of single extension steps with compatible signatures.
This process is conceptualized as a search over a graph where types are nodes and extension steps are edges.
For every binary operator $\genop$ with the signature $T \genop X : S$, an edge is created from type $T$ to type $S$.
As an example, the operator for numerical addition \ttt{+} has the signature $\ttt{number}\,\ttt{+}\,\ttt{number}\,:\,\ttt{number}$, thereby forming an edge from $\ttt{number}$ to itself.
Furthermore, for every member $\ttt{n}$ of type $T$, we create an edge from $T$ to $\lookup(T, \ttt{n})$, e.g., from $\ttt{number}$ to $\ttt{()}\ \ttt{=>}\ \ttt{string}$ for the member \ttt{toString} of \ttt{number} type.
Finally, we connect each function type $\ttt{(}\overline{p}\texttt{)}\ \ttt{=>}\ R$ and with its return type $R$.
For instance, $\ttt{()}\ \ttt{=>}\ \ttt{string}$ is connected with $\ttt{string}$.
Examples of type graphs can be found in \cref{sec:motivating-example,fig:intro-automaton-b}.
Note that these extension steps are \emph{abstract}, in the sense that they focus on the type of the expression being extended and the resulting type after extension, not considering textual representation and parameters.

The type reachability algorithm, \cref{alg:reachability}, implements a depth-first search over this type graph, starting from the current type $T$, succeeding upon finding goal type $G$ (\creflreach{alg:reachability:goal-found}), marking any visited types to prevent cycles (\creflreach{alg:reachability:cycle}).
Then, it proceeds to iterate over all valid extension steps from $T$ (\creflreach{alg:reachability:iter}) and computes the resulting type $S$ after the extension step is applied (\creflreach{alg:reachability:explore-next}).
In the conceptualized type graph, as described in the previous paragraph, this is equivalent to exploring all outgoing edges from $T$.
At \creflreach{alg:reachability:recurse-next}, we proceed to recursively search if $S$ can reach $G$.
If all recursive calls are unsuccessful, the goal type can not be reached (\creflreach{alg:reachability:not-found}).

\algnewcommand\algorithmicforeach{\textbf{for each}}
\algdef{S}[FOR]{ForEach}[1]{\algorithmicforeach\ #1\ \algorithmicdo}

\begin{algorithm}[t]
  \small
  \caption{Our type reachability search algorithm}
  \label{alg:reachability}
  \begin{algorithmic}[1]
  \Require Current type $T$ of some expression $e$, goal type $G$
  \Ensure Whether $G$ can be reached by extending $e$
  \Procedure{\reachable}{$T$, $G$}
  \State \textbf{if} $T = G$ \textbf{then} \textbf{return} true \label{alg:reachability:goal-found} \Comment{The goal type is successfully found}
  \State \textbf{if} $T$ is marked \textbf{then} \textbf{return} false \textbf{else} mark $T$ \label{alg:reachability:cycle} \Comment{Type $T$ is marked to avoid cycles}
  \ForEach{valid extension step $\lozenge$ from $T$} \label{alg:reachability:iter}
    \State $S \coloneq$ the resulting type of applying $\lozenge$ on $T$ \label{alg:reachability:explore-next}
    \State \textbf{if} $\textsc{pruneSearch}(T, G, S)$ \textbf{continue} \label{alg:reachability:filter-next} \Comment{Prune the search to ensure termination}
    \State \textbf{if} $\reachable(S, G)$ \textbf{return} true \label{alg:reachability:recurse-next} \Comment{Recurse to the next round of extension}
  \EndFor
  \State\textbf{return} false \label{alg:reachability:not-found} \Comment{No suitable extension is found}
  \EndProcedure
  \end{algorithmic}
\end{algorithm}

Some programming languages define self-referential default members, e.g., \ttt{clone} in Java or \ttt{valueOf} in TypeScript, which are nullary functions that return a value of the same type as the callee, $() => T$ for type $T$.
When these members are accessed in functions, higher-order functions can be derived indefinitely.
For instance, for a function $f$ with type $\ttt{()}\ \ttt{=>}\ S$, $f.\ttt{valueOf}$ has the type $\ttt{()}\ \ttt{=>}\ \ttt{()}\ \ttt{=>}\ S$.
We therefore need to restrict the type search to a finite set of types to ensure termination.
At \creflreach{alg:reachability:filter-next} of \cref{alg:reachability}, we add a heuristic $\textsc{pruneSearch}$ into the search, which decides where to prune the search process.
We develop a simple heuristic based on the results from \citet{completion-DBLP:conf/pldi/GveroKKP13}. This heuristic prunes exploration of types with higher complexity than goal or source type if they do not contain yet unexplored primitive types, thus preventing exploration of arbitrarily complex types.
The details of this heuristic are presented in \appcref{app:heuristic-reachable}.
While ensuring termination, our heuristic leads to incompleteness and the potential rejection of well-typed expressions.
However, this effect is less pronounced in practical usage, as only highly complex (thus less realistically used) types are avoided.

We proceed to prove the soundness of \cref{alg:reachability} below.

\begin{lemma}
  \label{lemma:reachable}
  The type search in \cref{alg:reachability} is \emph{sound}, i.e., for any expression $e$ with $\te \vdash e : T$, if $\reachable(T, G)$ holds, then there exists a sequence of extensions $y$ such that $\te \vdash e \strcat y : G$.
\end{lemma}
\begin{proof}
  By the design of \cref{alg:reachability}, if $\reachable(T, G)$ returns true, there is a sequence of $n$ recursive calls to $\reachable(T_i, G)$, with $T_0 = T$ and $\reachable(T_n, G) = \text{true}$.
  Each $T_i$ ($i > 0$) is derived because some extension $\lozenge_i$ is applicable to $T_{i-1}$ based on the typing rules of $L_B$.
  We then convert each $\lozenge_i$ to its concrete, textual version $\blacklozenge_i$. This representation includes the required well-typed parameters of $\blacklozenge_i$ (i.e., for binary operators and non-nullary functions), which are constructed using literals.
  Finally, we construct $y$ as $\blacklozenge_1 \strcat \dots \strcat \blacklozenge_n$.
\end{proof}

Note that using any pruning heuristic at \creflreach{alg:reachability:filter-next} of \cref{alg:reachability} preserves soundness, which in turn is sufficient to preserve the required prefix property, as defined in \cref{def:prefix}.
We can conclude that the two-tiered search algorithm soundly determines whether the desired target type can be derived from some partial input. Therefore, we conclude that $A_e \downarrow T$ and $A_e$ are prefix automata that parse a subset of well-typed expressions in $L_B$.

\begin{corollary}
  \label{cor:expr}
  If $\reachable(\der(\sstate), G)$ holds for any $\sstate \in \reachf(\initial_e, \istr)$ of a partial expression $\istr$, then there exists a suffix $\istr'$ such that $\te \vdash \istr \strcat \istr' : G$.
\end{corollary}

\begin{proof}
  This conclusion follows directly from \cref{lemma:derivable,lemma:reachable}.
\end{proof}

\begin{lemma}
The language parsed by $A_e \downarrow T$ is thus a subset of the expressions of $L_B$ of type $T$, i.e., $L(A_e \downarrow T) \subseteq \{\istr \mid \te \vdash \istr : T\}$.
Since $A_e$ recursively involves $A_e \downarrow T$, the language parsed by $A_e$ is also a subset of well-typed expressions of $L_B$, i.e., $L(A_e) \subseteq \{\istr \mid \exists T : \te \vdash \istr : T\}$.
\end{lemma}

\subsection{Prefix Automata for Statements}
\label{sec:ce-stmts}

We define the remaining automata to capture the complete language \subts{}.
The statement automaton is defined recursively as $A_s \coloneq A_{\textsc{decl}} \cup A_{\textsc{expr}} \cup A_{\textsc{ret}} \cup A_{\textsc{block}} \cup A_{\textsc{fun}} \cup A_{\textsc{ite}}$.
The declaration automaton $A_{\textsc{decl}} \coloneq A_{\ttt{let}\ x\, \ttt{:}\, T \ttt{;}}$ captures undefined variable names $x$ by accepting all strings, except for existing identifiers. This automaton is a prefix automaton since an accepting state can always be reached by appending characters to the declared identifier.
The return statement automaton is $A_{\varnothing}$ when outside a function and restricts the parsed expression to the return type of the surrounding function otherwise.
The remaining automata are mainly concatenations of previously defined automata and recursive invocations of $A_s$, with small variations detailed in \appcref{app:ce-statements-details}.

\paragraph{Tracking Type Environments}
Generally, we follow the typing rules in \cref{fig:inference-rules-stmt}. Identifiers are passed on through all state transitions, matching the rule \textsc{seq}, where the type environment of consecutive statements needs to be compatible.
However, in the cases of \textsc{block}, \textsc{ite} and \textsc{fun}, we discard the local type environment after parsing, matching the respective typing rules. In \textsc{fun} additionally, the function signature and parameters are added into the type environment of the function body automaton, and the function signature in the environment of subsequent statements.

\paragraph{Guaranteeing Return Types}
When parsing the body of a function, the transition function of the function automata $A_{\textsc{fun}}$ maintains information about the declared return type and the encountered return statements (if any).
$A_{\textsc{fun}}$ only accepts states where all return values match the declared return type and all execution paths inside the function body return, following $L_B$'s typing rules in \cref{fig:inference-rules-return}.
If the current generated statements do not return in all execution paths, another statement is forced to be generated.
Since we can always express the requested type through literals, a correct return statement can always be generated and the prefix automaton property is not violated.

The described rules are straightforward to implement without violating the prefix property as all restrictions are derived only from already parsed input, e.g., the already defined identifiers or the previously declared function return type. We can therefore deduce that the statement automaton is a prefix automaton. Moreover, the automaton accepts all valid statements of $\subts$, with the exception of well-typed expressions rejected by $A_e$. Therefore the parsed language is a subset of $\subts$.

\begin{lemma}
  With $A_{M} \coloneq A_{\overline{s}}$ it holds that $A_{M}$ is a prefix automaton and $L(A_{M}) \subseteq \subts$
\end{lemma}

%% file: figures/grammar/grammar-detailed.tex
\begin{figure}[tp]
  \small
  \centering
  \begin{minipage}{\textwidth}
    \begin{minipage}[t]{0.35\textwidth}
      \centering
      \begin{tabular}{l@{\hspace{5mm}}l}
$l ::= $                                                                                             & Literal                          \\
\quad $\ttt{\textbackslash{}d}+$                                                                     & \quad Numeric Literal            \\
\quad $\ttt{"}\ttt{\textbackslash{}w}*{"}$                                                           & \quad String Literal             \\
\quad $\ttt{true} \mid{} \ttt{false}$                                                                & \quad Boolean Literal            \\
\\
$x ::= \ttt{\textbackslash{}w}+$                                                                     & Identifier                       \\
\\
\revisionPC{$e ::= e_0 \mid{} e_1 $                                                                          }   &\revisionPC{ Expression                      } \\
\revisionPC{$e_0 ::= $                                                                                       }   &\revisionPC{ Base Expression                 } \\
\revisionPC{\quad $l$                                                                                        }   &\revisionPC{ \quad Literal                   } \\
\revisionPC{\quad $x$                                                                                        }   &\revisionPC{ \quad Identifier                } \\
\revisionPC{\quad $\ttt{(}{\overline{p}}\ttt{)}\ \ttt{=>}\ e$                                                }   &\revisionPC{ \quad Function Expression       } \\
\revisionPC{\quad $\ttt{(}e\ttt{)}$                                                                          }   &\revisionPC{ \quad Grouped Expression        } \\
\revisionPC{$e_1 ::= $                                                                                       }   &\revisionPC{ Extension Expression            } \\
\revisionPC{\quad $e \genop e$                                                                               }   &\revisionPC{ \quad Binary Operator           } \\
\revisionPC{\quad $e\ttt{(}{\overline{e}}\ttt{)}$                                                            }   &\revisionPC{ \quad Function Call             } \\
\revisionPC{\quad $e\ttt{.}n$                                                                                }   &\revisionPC{ \quad Member Access             } \\
      \end{tabular}
    \end{minipage}
    \hspace{5mm}
    \begin{minipage}[t]{0.01\textwidth}
      \begin{tabular}{l|l}
& \\
& \\
& \\
& \\
& \\
& \\
& \\
& \\
& \\
& \\
& \\
& \\
& \\
& \\
& \\
& \\
& \\
      \end{tabular}
    \end{minipage}
    \hspace{3mm}
    \begin{minipage}[t]{0.25\textwidth}
      \centering
      \begin{tabular}{l@{\hspace{4mm}}l}
$p ::= x\, \ttt{:}\, T$                                                                              & Typed Identifier                 \\
\\
$T ::= $                                                                                             & Type                             \\
\quad $\ttt{number}$                                                                                 & \quad Numeric Type               \\
\quad $\ttt{string}$                                                                                 & \quad String Type                \\
\quad $\ttt{boolean}$                                                                                & \quad Boolean Type               \\
\quad $\ttt{(}\overline{p}\texttt{)}\ \ttt{=>}\ T$                                                   & \quad Function Type              \\
\\
$s ::= $                                                                                             & Statement                        \\
\quad $\ttt{let}\ x\, \ttt{:}\, T\ttt{;}$                                                            & \quad Variable Declaration \\
\quad $e\ttt{;}$                                                                                     & \quad Expression Statement       \\
\quad $\ttt{return}\ e\ttt{;}$                                                                       & \quad Return Statement           \\
\quad $\ttt{\{}\,\overline{s}\,\ttt{\}}$                                                                 & \quad Statement Block            \\
\quad $\ttt{function}\ x\,\ttt{(}{\overline{p}}\ttt{)}\, \ttt{:}\, T\, \ttt{\{}\,\overline{s}\,\ttt{\}}$ & \quad Function Definition        \\
\quad $\ttt{if}\ \ttt{(}e\ttt{)}\ s\ \ttt{else}\ s$                                                  & \quad If-Then-Else Statement     \\
\\
$M ::= {\overline{s}}$                                                                               & Program                          \\
      \end{tabular}
    \end{minipage}
  \end{minipage}
  \vspace{-2mm}
  \caption{The syntax of \subts{}. Expressions are categorized into base and extension expressions. The later extends a given expression with suffix operators to form more complicated expressions.}
  \label{fig:subset-l-detailed}
\end{figure}

%% file: figures/grammar/inference-expr.tex
\begin{figure}
  \small
  \centering
      \begin{equation*}
      \begin{gathered}
    [\textsc{lit-num}]\inference{}{\te |- \ttt{\textbackslash{}d+} : \texttt{number}}
    \quad
    [\textsc{lit-str}]\inference{}{\te |- \texttt{"\textbackslash{}w*"}: \texttt{string}}
    \quad
    [\textsc{lit-bool}]\inference{}{\te |- \texttt{true},\texttt{false} : \texttt{boolean}}
    \\[2mm]
    [\textsc{ident}]\inference{(x: T) \in \te}{\te |- x : T}
    \quad
    \revisionPC{[\textsc{anon}]\inference{\te \cup \overline{p} |- e : T}{\te |- \ttt{(}{\overline{p}}\ttt{)}\ \ttt{=>}\ e : (\overline{p}) \ttt{ => } T}}
    \quad
    \revisionPC{[\textsc{call}]\inference{\te |- f : (\overline{x}: \overline{S}) \ttt{ => } T \quad \te |- \overline{e} : \overline{S}}{\te |- f\ttt{(}\overline{e}\ttt{)} : T}}
    \\[2mm]
    [\textsc{group}]\inference{\te |- e : T }{\te |- \ttt{(}e\ttt{)} : T}
    \ \,
    [\textsc{op}]\inference{\te |- e_1 : S_1 \quad \te |- e_2 : S_2 \quad S_1 \genop S_2 : T}{\te |- e_1 \genop e_2 : T}
    \ \,
    [\textsc{mem}]\inference{\te |- e : S \quad \lookup(S, n) = T}{\te |- e\ttt{.}n : T}
      \end{gathered}
      \end{equation*}
    \vspace{-2mm}
    \caption{Typing rules for \subts{}'s expressions.}
  \label{fig:inference-rules-expr}
\end{figure}

%% file: figures/grammar/inference-stmt.tex
\begin{figure}
  \small
  \centering
  \begin{tabular}{@{}p{0.8\textwidth}@{}}
    \begin{equation*}
      \begin{gathered}
    [\textsc{decl}]\inference{x \notin \te}{\te |- \ttt{let}\ x\, \ttt{:}\, T \ttt{;} \ext \te \cup (x:T)}
    \quad
    [\textsc{expr}]\inference{\te |- e : T}{\te |- e\ttt{;} \ext \te}
    \quad
    [\textsc{ret}]\inference{\text{inside function body} \quad \te |- e : T}{\te |- \ttt{return}\ e\ttt{;} \ext \te}
    \\[2mm]
    [\textsc{block}]\inference{\te |- \overline{s_B} \ext \te_B}{\te |- \ttt{\{}\,\overline{s_B}\,\ttt{\}} \ext \te}
    \quad
    [\textsc{fun}]\inference{x \notin \te \quad \quad \te \cup (x: (\overline{p}) \ttt{ => } T) \cup (\overline{p}) |- \overline{s_x} \ext \te_{x}}{\te_1 |- \ttt{function}\ x\, \ttt{(}\overline{p}\ttt{)}\, \ttt{:}\, T\, \ttt{\{}\,\overline{s_x}\,\ttt{\}} \ext \te \cup (x: (\overline{p}) \ttt{ => } T)}
    \\[2mm]
    [\textsc{ite}]\inference{\te |- s_{if} \ext \te_{if} \quad \te |- s_{else} \ext \te_{else}}{\te |- \ttt{if}\ \ttt{(} e\ttt{)}\ s_{if}\ \ttt{else}\ s_{else} \ext \te}
    \quad
    [\textsc{nop}]\inference{}{\te |- \bullet \ext \te}
    \quad
    [\textsc{seq}]\inference{\te_1 |- \overline{s} \ext \te_2 \quad \te_2 |- s \ext \te_3}{\te_1 |- \overline{s}~s \ext \te_3}
      \end{gathered}
    \end{equation*}
  \end{tabular}
  \vspace{-4mm}
    \caption{Type environment extension rules for sequences of statements in \subts{}.}
  \label{fig:inference-rules-stmt}
\end{figure}

%% file: figures/grammar/inference-ret.tex
\begin{figure}
  \small
  \centering
  \begin{tabular}{@{}p{0.8\textwidth}@{}}
    \begin{gather*}
    [\textsc{r-decl}]\inference{\te |- \overline{s} : R }{\te |- \ttt{let}\ x\, \ttt{:}\, T\ttt{;} \overline{s} : R}
    \quad
    [\textsc{r-expr}]\inference{\te |- \overline{s} : R }{\te |- e\ttt{;} \overline{s} : R}
    \quad
    [\textsc{r-ret}]\inference{\te |- e : R}{\te |- \ttt{return}\ e\ttt{;} \overline{s} : R}
    \\[2mm]
    [\textsc{r-block-self}]\inference{\te |- \overline{s_B} : R \quad \te |- \overline{s} }{\te |- \ttt{\{}\, \overline{s_B}\, \ttt{\}}\, \overline{s} : R}
    \quad
    [\textsc{r-block-next}]\inference{\te |- \overline{s_B} \quad \te |- \overline{s} : R }{\te |- \ttt{\{}\, \overline{s_B}\, \ttt{\}}\, \overline{s} : R}
    \\[2mm]
    [\textsc{r-fun}]\inference{\te \cup (x: (\overline{p} \ttt{ => } R)) |- \overline{s} : R' \quad \te \cup (x: (\overline{p}) \ttt{ => } R) \cup (\overline{p}) |- \overline{s_x} : R}{\te |- \ttt{function}\ x\,\ttt{(}\overline{p}\ttt{)}\, \ttt{:}\, R\, \ttt{\{}\, \overline{s_x}\, \ttt{\}}\; \overline{s} : R'}
    \\[2mm]
    [\textsc{r-ite-self}]\inference{\te |- s_{if} : R \quad \te |- s_{else} : R}{\te |- \ttt{if}\ \ttt{(} e \ttt{)}\ s_{if}\ \ttt{else}\ s_{else}\ \overline{s} : R}
    \quad
    [\textsc{r-ite-next}]\inference{\te |- \overline{s} : R}{\te |- \ttt{if}\ \ttt{(} e \ttt{)}\ s_{if}\ \ttt{else}\ s_{else}\ \overline{s} : R}
    \end{gather*}
  \end{tabular}
  \vspace{-4mm}
    \caption{\subts{}'s typing rules for function returns.}
  \label{fig:inference-rules-return}
\end{figure}

%% file: sections/typescript.tex
\section{Extension to TypeScript}
\label{sec:typescript}

We extend our completion engine described in \cref{sec:method} to handle a core subset of modern TypeScript.
In this section, we selectively discuss the implementation of several interesting TypeScript features.
We provide a comprehensive list of supported and unsupported TypeScript features in \appcref{app:experiments-details}.

\paragraph{Constant Variable Declarations}
In addition to variable declaration using \ttt{let}, TypeScript supports constant declarations using \ttt{const}. This defines immutable identifiers.
We thus additionally track mutability of each identifier in the type environment and disallow applying the assignment operator to immutable identifiers.

\paragraph{Arrays}
We add support for array type annotation, parsing array expressions, and reading from and assigning to array fields.
In array expressions, we enforce that all array elements have the same type.
Moreover, array types introduce another dimension of type nesting. Therefore we adapt the type reachability pruning heuristic to handle this additional dimension to ensure termination.

\paragraph{Loops}
TypeScript supports various loop constructs, including \texttt{for}, \texttt{while}, \texttt{do-while}, and \texttt{for..of} loops. These are implemented mostly as variations of the statement block parser. The \texttt{for..of} loop uniquely constrains the right-hand side of the \texttt{..of} operator to an array of any type. To adapt the type search, we introduce a generic array type $\bullet[]$, which matches any array type. For example, both types \ttt{number[]} and \ttt{string[]} match $\bullet[]$ in \creflreach{alg:reachability:goal-found} of \cref{alg:reachability}.

\paragraph{Additional Operators and Types}
We add several arithmetic and logic operators, such as modulo \ttt{\%}, exact equality \ttt{===}, logical or \ttt{||}, and the ternary operator \ttt{?:}.
To handle these operators, we add additional edges to the type search graph.
Moreover, we add support for post- and prefix operators such as \ttt{---} and \ttt{++}, which are only valid extensions to mutable expressions.

\paragraph{Operator Precedence}
TypeScript defines an operator precedence, which determines the implicit grouping of expressions.
For example \ttt{1 + 2.toString()} is parsed as \ttt{1 + (2.toString())}.
We adapt our expression parsing algorithm in two places to handle operator precedences. First, in the expression automaton, we leverage the knowledge about previously parsed extensions to determine the implicit grouping and thus where the next operator is applied.
For example, for state $q^{\ttt{1 + 2}}$, the member access extension $\ttt{.}n$ is applied to \ttt{2}, as opposed to \ttt{1 + 2}.
Second, we adapt the type search in \cref{alg:reachability}. Concretely, we ensure that only extensions that can be validly applied based on operator precedence are iterated over. For this, we track the operator precedence of previously parsed extensions and extensions considered during the traversal of the type graph and omit operators in \creflreach{alg:reachability:explore-next} that violate operator precedence.

\paragraph{Global Identifiers and Imports}
In TypeScript, many identifiers are defined globally and available in any execution. These global identifiers are incorporated by initializing the type environment of the program automaton accordingly. Identifiers such as \texttt{Math} introduce additional types, which we additionally implement. We also model the import of the \ttt{crypto} library using \ttt{require}.

\paragraph{Polymorphic Built-In Members}
The TypeScript \lookup{} implementation defines a few polymorphic members for built-in types. For example, for array \ttt{x} of type \texttt{T[]}, \texttt{x.map(f)} takes a callback function \ttt{f} and returns a new array \ttt{[f(x[0]), f(x[1]), ...]}. If \ttt{f} has type \ttt{(T) => P}, the returned array has type \ttt{P[]}. Here \texttt{P} is a \emph{type parameter}, which is instantiated by matching the type of the passed function to the type pattern.

We support such polymorphisms by adapting the type search. We track type patterns and enforce that type parameters are instantiated before the goal type is reached. We then continue the search from the instantiated version. In the \ttt{map} example, when searching completions of \ttt{x.map}, we first search for functions that instantiate the type parameter, and then continue the search from the instantiated type. When anonymous functions are generated as call parameters, we enforce that the function matches the searched type pattern.

\paragraph{Type Annotations}
TypeScript is designed to be flexible, allowing many type annotations to be omitted when they can be automatically inferred. We generally support this, such as inferring types from initial values. However, it can lead to unexpected types when annotations are omitted, often confusing even experienced developers \citep{typescript-issue-59999,typescript-issue-59863}.
Moreover, in the context of LLM-based code generation, having more type annotations can provide valuable information for both the model and our type-constraining algorithms.
We have identified three situations where generated code often fails to compile without type annotations, prompting us to enforce them. First, we require annotations for all function parameters and return types. Second, all variable declarations must either have a type annotation or be initialized with a value. Third, we enforce type annotations for the first parameter of anonymous functions used as callbacks in the polymorphic built-in member \ttt{reduce}. These constraints trade off practical correctness with theoretical language completeness.

%% file: sections/experiments.tex
\section{Experimental Evaluation}
\label{sec:experiments}

We present an extensive evaluation of our type constraining approach on a variety of tasks and models. We outline our experimental setup (\cref{sec:experiments-setup}), evaluate the impact on compilation errors and functional correctness (\cref{sec:experiments-results}), perform runtime analysis (\cref{sec:experiments-runtime}), and present case studies (\cref{sec:case-study}).

\subsection{Experimental Setup}
\label{sec:experiments-setup}

We now outline our main evaluation setup, covering implementation, evaluated tasks, considered models, compared methods, and metrics. We provide further setup details and hyperparameter choices in \appcref{app:experiments-details}.

\revisionPC{
\paragraph{Implementation}
Our implementation is written in Python and contains 11249 lines of code.
To ensure robust implementation, we built a large set of around four hundred unit tests and frequently compared the behaviors of our implementation with the official TypeScript compiler~\citep{tsc}.
}

\paragraph{Tasks and Benchmarks}
We evaluate three relevant tasks of code generation:
\begin{itemize}[leftmargin=7mm]
\item \emph{Synthesis}: Given a natural language task description and a function header, the task is to generate a solution from scratch.
\item \emph{Translation}: Given a function written in Python and the header of an equivalent TypeScript function, the task is to generate the body of the equivalent function in TypeScript.
\item \emph{Repair}: Given a natural language task description, a non-compilable solution, the corresponding compiler error, and the function header, the task is to restore functionality of the flawed solution by resolving the compilation error.
\end{itemize}
The benchmarks for these tasks are based on TypeScript-translated tasks from HumanEval \citep{few-shot-learners-DBLP:conf/nips/BrownMRSKDNSSAA20} and MBPP \citep{austin2021programsynthesislargelanguage-mbpp}, contained in the MultiPL-E dataset \citep{multiple-DBLP:journals/tse/CassanoGNNPPYZAFGGJ23}, with 159 and 384 instances each. We observe that success in generating valid code for the same sample can vary depending on the random seed used. To obtain more comprehensive results on the small HumanEval dataset, we generate each sample $4$ times with different seeds and aggregate the outcomes. In MBPP, we generate each sample once.
For Repair, we collect all non-compiling programs from the unconstrained synthesis task for all models, resulting in 292 and 248 instances for HumanEval and MBPP each.

\paragraph{Models}
We use 6 different open-weight LLMs, covering 3 LLMs of varying parameter sizes from the same model family and 4 models of a similar size from different model families: the Gemma 2 model family with 2B/9B/27B parameters \citep{gemma-2024}, DeepSeek Coder 33B (abbreviated as DS Coder 33B) \citep{deepseek-coder-DBLP:journals/corr/abs-2401-14196}, CodeLlama 34B \citep{code-llama-DBLP:journals/corr/abs-2308-12950}, and Qwen2.5 32B \citep{qwen2025qwen25technicalreport}. For all evaluated LLMs we choose the instruction-tuned variants, which are fine-tuned to follow instructions in a chat-style interaction, such that they adequately attempt to resolve the presented tasks.

\paragraph{Compared Methods}
We run unconstrained LLM sampling, reported as \emph{Vanilla}.
We measure the upper bound improvement of prior syntactic constraining methods~\citep{synchromesh,ugare2024syncode,dominos-DBLP:journals/corr/abs-2403-06988} by assuming that all syntactically incorrect instances generated by Vanilla could be compiled under syntactic constraining.
We refer to this improvement as idealized \emph{Syntax}.
We separately sample using type-constrained decoding based on our completion engine introduced in \cref{sec:method,sec:typescript}, and report it as \emph{Types}.
Due to the size and complexity of the full TypeScript compiler, featuring over 427,105 lines of code in 698 files \citep{tsc}, our extension does not cover all features of TypeScript.
We therefore emulate a type constraining that supports the entire TypeScript feature set. Concretely, if a sample compiles correctly without any constraining, we report it as-is. Otherwise, we report the result of a constrained resample.
For all methods, if generation takes more than \SI{300}{seconds}, we report the partial program generated until the timeout.

\revisionRB{
\paragraph{Metrics}
We compute two main metrics to assess the effectiveness of the compared methods. First, we determine the number of \emph{compiler errors} in model-generated outputs.
We count as a compiler error any case in which the TypeScript compiler \citep{tsc} reports an issue during compilation. 
To measure \emph{functional correctness}, we leverage the pass@1 metric \citep{humaneval}, which measures the percentage of code generations that pass the provided unit tests given only one trial.
}

\begin{table}
    \renewcommand{\arraystretch}{1.0}
    \setlength{\tabcolsep}{3pt}
    \caption{\revisionOC{Number of instances with compiler errors in unconstrained generation (Vanilla), idealized syntax-only constraining (Syntax), and our proposed type constraining (Types). Type constraining reduces compiler errors by 74.8\% and 56.0\% in the synthesis of HumanEval and MBPP problems respectively, compared to only 9.0\% and 4.8\% ideal improvement on the two datasets respectively through syntax-only constraining.} }
    \vspace{-3mm}
    \small
    \begin{tabular}{@{}ll@{} cr@{}lr@{}r@{} p{.5mm} cr@{}lr@{}r@{} p{.5mm} cr@{}lr@{}r@{}}
        \toprule
        && \multicolumn{5}{c}{Synthesis} && \multicolumn{5}{c}{Translation} && \multicolumn{5}{c}{Repair} \\
        \cmidrule{3-7} \cmidrule{9-13} \cmidrule{15-19}
        
        & Model & Vanilla & \multicolumn{2}{c}{Syntax} & \multicolumn{2}{r}{Types} && Vanilla & \multicolumn{2}{c}{Syntax} & \multicolumn{2}{r}{Types} && Vanilla & \multicolumn{2}{c}{Syntax} & \multicolumn{2}{r}{Types} \\
        \midrule
        \parbox[t]{3mm}{\multirow{6}{*}{\rotatebox[origin=c]{90}{HumanEval}}}
        & Gemma 2 2B    & $103$      & $92$&$_{\downarrow 10.7\%}$ & $\textbf{44}$&$_{\downarrow 57.3\%}$ &    & $177$      & $149$&$_{\downarrow 15.8\%}$ & $\textbf{80}$&$_{\downarrow 54.8\%}$ &    & $194$      & $181$&$_{\downarrow 6.7\%}$ & $\textbf{103}$&$_{\downarrow 46.9\%}$ \\
        & Gemma 2 9B    & $45$       & $41$&$_{\downarrow 8.9\%}$  & $\textbf{13}$&$_{\downarrow 71.1\%}$ &    & $75$       & $63$&$_{\downarrow 16.0\%}$  & $\textbf{16}$&$_{\downarrow 78.7\%}$ &    & $113$      & $108$&$_{\downarrow 4.4\%}$ & $\textbf{52}$&$_{\downarrow 54.0\%}$  \\
        & Gemma 2 27B   & $15$       & $13$&$_{\downarrow 13.3\%}$ & $\textbf{2}$&$_{\downarrow 86.7\%}$  &    & $20$       & $20$&$_{\downarrow 0.0\%}$   & $\textbf{3}$&$_{\downarrow 85.0\%}$  &    & $45$       & $40$&$_{\downarrow 11.1\%}$ & $\textbf{22}$&$_{\downarrow 51.1\%}$  \\
        & DS Coder 33B  & $26$       & $25$&$_{\downarrow 3.8\%}$  & $\textbf{5}$&$_{\downarrow 80.8\%}$  &    & $18$       & $17$&$_{\downarrow 5.6\%}$   & $\textbf{7}$&$_{\downarrow 61.1\%}$  &    & $36$       & $36$&$_{\downarrow 0.0\%}$  & $\textbf{15}$&$_{\downarrow 58.3\%}$  \\
        & CodeLlama 34B & $86$       & $71$&$_{\downarrow 17.4\%}$ & $\textbf{28}$&$_{\downarrow 67.4\%}$ &    & $158$      & $124$&$_{\downarrow 21.5\%}$ & $\textbf{59}$&$_{\downarrow 62.7\%}$ &    & $153$      & $142$&$_{\downarrow 7.2\%}$ & $\textbf{48}$&$_{\downarrow 68.6\%}$  \\
        & Qwen2.5 32B   & $17$       & $17$&$_{\downarrow 0.0\%}$  & $\textbf{2}$&$_{\downarrow 88.2\%}$  &    & $24$       & $21$&$_{\downarrow 12.5\%}$  & $\textbf{5}$&$_{\downarrow 79.2\%}$  &    & $36$       & $34$&$_{\downarrow 5.6\%}$  & $\textbf{13}$&$_{\downarrow 63.9\%}$  \\
        \noalign{\vspace{0.7mm}}
        \midrule
        \parbox[t]{3mm}{\multirow{6}{*}{\rotatebox[origin=c]{90}{MBPP}}}
        & Gemma 2 2B    & $67$       & $64$&$_{\downarrow 4.5\%}$  & $\textbf{27}$&$_{\downarrow 59.7\%}$ &    & $126$      & $111$&$_{\downarrow 11.9\%}$ & $\textbf{79}$&$_{\downarrow 37.3\%}$ &    & $194$      & $184$&$_{\downarrow 5.2\%}$ & $\textbf{108}$&$_{\downarrow 44.3\%}$ \\
        & Gemma 2 9B    & $30$       & $29$&$_{\downarrow 3.3\%}$  & $\textbf{10}$&$_{\downarrow 66.7\%}$ &    & $67$       & $61$&$_{\downarrow 9.0\%}$   & $\textbf{33}$&$_{\downarrow 50.7\%}$ &    & $129$      & $124$&$_{\downarrow 3.9\%}$ & $\textbf{63}$&$_{\downarrow 51.2\%}$  \\
        & Gemma 2 27B   & $20$       & $19$&$_{\downarrow 5.0\%}$  & $\textbf{7}$&$_{\downarrow 65.0\%}$  &    & $37$       & $36$&$_{\downarrow 2.7\%}$   & $\textbf{22}$&$_{\downarrow 40.5\%}$ &    & $71$       & $69$&$_{\downarrow 2.8\%}$  & $\textbf{32}$&$_{\downarrow 54.9\%}$  \\
        & DS Coder 33B  & $32$       & $32$&$_{\downarrow 0.0\%}$  & $\textbf{19}$&$_{\downarrow 40.6\%}$ &    & $29$       & $27$&$_{\downarrow 6.9\%}$   & $\textbf{13}$&$_{\downarrow 55.2\%}$ &    & $90$       & $90$&$_{\downarrow 0.0\%}$  & $\textbf{43}$&$_{\downarrow 52.2\%}$  \\
        & CodeLlama 34B & $80$       & $71$&$_{\downarrow 11.2\%}$ & $\textbf{41}$&$_{\downarrow 48.8\%}$ &    & $126$      & $114$&$_{\downarrow 9.5\%}$  & $\textbf{54}$&$_{\downarrow 57.1\%}$ &    & $157$      & $148$&$_{\downarrow 5.7\%}$ & $\textbf{76}$&$_{\downarrow 51.6\%}$  \\
        & Qwen2.5 32B   & $19$       & $18$&$_{\downarrow 5.3\%}$  & $\textbf{13}$&$_{\downarrow 31.6\%}$ &    & $22$       & $22$&$_{\downarrow 0.0\%}$   & $\textbf{16}$&$_{\downarrow 27.3\%}$ &    & $55$       & $52$&$_{\downarrow 5.5\%}$  & $\textbf{29}$&$_{\downarrow 47.3\%}$  \\
        \bottomrule
        \end{tabular}
        \label{fig:main-synth-translation-res}
        \label{fig:repair-res}
\end{table}

\subsection{Results on Compilation and Functional Correctness}
\label{sec:experiments-results}

In this section, we present our experimental results, showing that on all three code-generation-related tasks, our type constraining approach significantly improves the considered LLMs in generating both compilable and functionally correct code. It also substantially outperforms syntax-only constraining.

\paragraph{Reduction of Compilation Errors}
In \cref{fig:main-synth-translation-res}, we present the number of compilation errors produced by each compared method.
For synthesis and translation, in the unconstrained setting (Vanilla), on average only $9.0\%$ and $4.9\%$ of the non-compiling instances in HumanEval and MBPP respectively are due to syntactic errors (Syntax), with Qwen2.5 32B even making no syntax errors at all for HumanEval synthesis and MBPP translation.
In contrast, type constraining reduces compilation errors by more than half, i.e., by $75.3\%$ and $52.1\%$ on HumanEval and MBPP respectively.
We observe that models across all sizes and families benefit similarly from our constraining, with a minimum error reduction of $54.8\%$ and $27.3\%$ on HumanEval and MBPP respectively, highlighting the general effectiveness of our approach.

A straightforward way to improve successful compilation of LLM-generated code is to feed the erroneous code and the error message back to an LLM for correction --- our repair task.
Thanks to its general applicability, our type constraining approach can also enhance this process.
Our experimental results in the setting of code repair are also depicted in \cref{fig:repair-res}. 
We find that, in the vanilla setting, many models struggle to correctly localize and resolve compilation errors, with Gemma 2 2B for example repairing only $33.5\%$ and $25.8\%$ of the non-compiling HumanEval and MBPP instances, respectively.
This is substantially increased to $56.4\%$ and $58.4\%$ through type constraining.
On average, using type-constrained sampling, $53.7\%$ more compilation errors are resolved than using vanilla LLM decoding.


\begin{table}
  \centering
  \setlength{\tabcolsep}{8pt}
  \renewcommand{\arraystretch}{1.0}
  \caption{\revisionOC{pass@1 of unconstrained generation (Vanilla) and type constraining (Types). The benefit of our type-constraining approach transfers from reduced compilation errors to improved functional correctness.}}
  \label{fig:fc-result}
  \vspace{-3mm}
  \small
  \begin{tabular}{@{}ll@{} p{8mm} @{}cr@{}l@{} p{8mm} @{}cr@{}l@{} p{8mm} @{}cr@{}l@{}}
      \toprule
      &&& \multicolumn{3}{c}{Synthesis} && \multicolumn{3}{c}{Translation} && \multicolumn{3}{c}{Repair}  \\
      \cmidrule{4-6} \cmidrule{8-10} \cmidrule{12-14}
      & Model && Vanilla & \multicolumn{2}{r}{Types} && Vanilla & \multicolumn{2}{r}{Types} && Vanilla & \multicolumn{2}{r}{Types} \\
      \midrule
      \parbox[t]{3mm}{\multirow{6}{*}{\rotatebox[origin=c]{90}{HumanEval}}}
      & Gemma 2 2B    &            & $29.1$  & $\mathbf{30.2}$&$_{\uparrow 3.8\%}$ &            & $50.2$  & $\mathbf{53.9}$&$_{\uparrow 7.5\%}$ &            & $11.6$  & $\mathbf{20.9}$&$_{\uparrow 79.4\%}$ \\
      & Gemma 2 9B    &            & $56.6$  & $\mathbf{58.3}$&$_{\uparrow 3.1\%}$ &            & $73.7$  & $\mathbf{78.3}$&$_{\uparrow 6.2\%}$ &            & $24.0$  & $\mathbf{34.9}$&$_{\uparrow 45.7\%}$ \\
      & Gemma 2 27B   &            & $69.5$  & $\mathbf{71.2}$&$_{\uparrow 2.5\%}$ &            & $86.6$  & $\mathbf{87.7}$&$_{\uparrow 1.3\%}$ &            & $38.4$  & $\mathbf{41.1}$&$_{\uparrow 7.1\%}$  \\
      & DS Coder 33B  &            & $68.9$  & $\mathbf{71.1}$&$_{\uparrow 3.2\%}$ &            & $88.7$  & $\mathbf{90.1}$&$_{\uparrow 1.6\%}$ &            & $47.6$  & $\mathbf{50.7}$&$_{\uparrow 6.5\%}$  \\
      & CodeLlama 34B &            & $41.0$  & $\mathbf{43.4}$&$_{\uparrow 5.7\%}$ &            & $58.6$  & $\mathbf{63.5}$&$_{\uparrow 8.3\%}$ &            & $17.5$  & $\mathbf{27.4}$&$_{\uparrow 56.9\%}$ \\
      & Qwen2.5 32B   &            & $79.6$  & $\mathbf{81.8}$&$_{\uparrow 2.8\%}$ &            & $92.1$  & $\mathbf{93.9}$&$_{\uparrow 1.9\%}$ &            & $65.4$  & $\mathbf{71.2}$&$_{\uparrow 8.9\%}$  \\
      \midrule
      \parbox[t]{3mm}{\multirow{6}{*}{\rotatebox[origin=c]{90}{MBPP}}}
      & Gemma 2 2B    &            & $40.4$  & $\mathbf{42.4}$&$_{\uparrow 5.2\%}$ &            & $52.3$  & $\mathbf{56.0}$&$_{\uparrow 7.0\%}$  &            & $12.1$  & $\mathbf{22.6}$&$_{\uparrow 86.7\%}$ \\
      & Gemma 2 9B    &            & $65.4$  & $\mathbf{67.4}$&$_{\uparrow 3.2\%}$ &            & $71.4$  & $\mathbf{75.8}$&$_{\uparrow 6.2\%}$  &            & $24.2$  & $\mathbf{31.9}$&$_{\uparrow 31.7\%}$ \\
      & Gemma 2 27B   &            & $70.6$  & $\mathbf{72.1}$&$_{\uparrow 2.2\%}$ &            & $83.1$  & $\mathbf{84.4}$&$_{\uparrow 1.6\%}$  &            & $39.1$  & $\mathbf{45.2}$&$_{\uparrow 15.5\%}$ \\
      & DS Coder 33B  &            & $65.4$  & $\mathbf{67.2}$&$_{\uparrow 2.8\%}$ &            & $85.9$  & $\mathbf{89.1}$&$_{\uparrow 3.6\%}$  &            & $35.1$  & $\mathbf{43.1}$&$_{\uparrow 23.0\%}$ \\
      & CodeLlama 34B &            & $42.2$  & $\mathbf{45.6}$&$_{\uparrow 8.0\%}$ &            & $55.7$  & $\mathbf{63.3}$&$_{\uparrow 13.6\%}$ &            & $15.7$  & $\mathbf{26.6}$&$_{\uparrow 69.2\%}$ \\
      & Qwen2.5 32B   &            & $76.3$  & $\mathbf{76.6}$&$_{\uparrow 0.3\%}$ &            & $89.6$  & $\mathbf{90.4}$&$_{\uparrow 0.9\%}$  &            & $48.0$  & $\mathbf{54.0}$&$_{\uparrow 12.6\%}$ \\
      \bottomrule
      \end{tabular}
\end{table}

\paragraph{Improving Functional Correctness}
Programs that do not compile are always functionally incorrect.
With our type constraining method, non-compilable generations can be turned into well-formed ones, offering the possibility of achieving functional correctness.
In \cref{fig:fc-result}, we experimentally show that type constraining universally improves the functional correctness of LLM-generated code.
On the three tasks considered, employing type constraining improves LLMs' pass@1 rate, achieving an average increase by $3.5\%$ in synthesis, $5.0\%$ in translation, and $37.0\%$ in repair tasks. The larger improvement in the latter is due to vanilla LLMs generally struggling to generate functionally correct code.
One interesting phenomenon is that, for stronger models,
\begin{wraptable}[11]{r}{0.4\textwidth}
  \small
  \centering
  \vspace{-4mm}
  \renewcommand{\arraystretch}{1.0}
  \caption{Median time per synthesis instance in seconds spent by our type-constrained decoding and its relative increase compared with unconstrained decoding (Vanilla).}
  \label{fig:time-increase}
  \vspace{-3mm}
  \begin{tabular}{@{}lr@{}lr@{}r@{}}
    \toprule
    Model                  & \multicolumn{2}{c}{HumanEval}                    & \multicolumn{2}{r}{MBPP}     \\
    \midrule
    Gemma 2 2B             & $6.7$&$_{\uparrow 38.3\%}$   & $6.3$&$_{\uparrow 35.4\%}$  \\
    Gemma 2 9B             & $8.3$&$_{\uparrow 29.2\%}$   & $9.5$&$_{\uparrow 46.8\%}$  \\
    Gemma 2 27B            & $11.7$&$_{\uparrow 19.9\%}$  & $11.7$&$_{\uparrow 32.8\%}$ \\
    DS Coder 33B           & $11.5$& $_{\uparrow 36.2\%}$ & $9.4$&$_{\uparrow 59.5\%}$  \\
    CodeLlama 34B          & $7.6$&$_{\uparrow 40.8\%}$   & $7.0$&$_{\uparrow 37.6\%}$  \\
    Qwen2.5 32B            & $7.3$&$_{\uparrow 39.6\%}$   & $4.9$&$_{\uparrow 54.8\%}$  \\
    \bottomrule
  \end{tabular}
\end{wraptable}
constraints more likely lead to recovering functionally correct code. For example on the synthesis task, for Gemma 2 27B, out of the $26$ instances that required resampling to compile successfully, $17$ are also functionally correct. For Qwen2.5 32B, $15$ out of $21$ such instances were correct.


\subsection{\revisionOC{Runtime Analysis}}
\label{sec:experiments-runtime}

As discussed in \cref{sec:background}, compared with vanilla LLM decoding, our constrained decoding algorithm runs an additional loop (\crefldecoding{line:decoding-loop2} of \cref{alg:decoding}), where tokens are sampled from an LLM-produced next-token probability distribution and checked against the completion engine.
In this section, we investigate how this process introduces additional runtime overhead for our type constraining.
\revisionPC{Note that for each selected token, vanilla and constrained decoding both run LLM inference only once, meaning that there is no extra overhead from LLM inference in constrained decoding.}

\paragraph{Overhead of Type Constraining}
For an application of our method in practice, the effective runtime increase due to constrained decoding is highly relevant.
To assess it, we measure the runtime per synthesis instance in HumanEval and MBPP for both unconstrained and type-constrained decoding. We report the median runtime per instance for type constraining and its relative increase to unconstrained decoding in \cref{fig:time-increase}.
On average over the evaluated models, we observe a relative increase of $39.1\%$ and $52.1\%$ in HumanEval and MBPP respectively.
We consider this impact to be bearable for the observed significant decrease in compilation errors.
Moreover, this is measured on an unoptimized, Python-based implementation and could be significantly improved by a more system-oriented implementation, such as the one proposed by \citet{DBLP:journals/corr/abs-2411-15100}.

\begin{wrapfigure}[14]{r}{0.4\textwidth}
  \centering
  \vspace{-3mm}
  \resizebox{0.4\textwidth}{!}{
    \begin{tikzpicture}
      \begin{axis}[
        ybar,
        width=7cm,
        height=4.8cm,
        ylabel={\footnotesize{Count}},
        y label style={at={(axis description cs:0.12,.5)},anchor=south},
        xlabel={\footnotesize{Number of loop iterations}},
        x label style={at={(axis description cs:0.5,-0.05)},anchor=north},
        xtick pos=bottom,
        ytick pos=left,
        xmin=0.5,
        xmax=150000,
        log basis y=10,
        ymode=log,
        xmode=log,
        ymin=1,
        ymax=500000,
        bar width=0.025cm,
        axis lines=left,
        xtick align=outside,
        ytick align=inside,
        tick label style={font=\footnotesize}
      ]
        \addplot[style={fill=myblue, draw=myblue}] coordinates {(1, 86394) (2, 362) (3, 81) (4, 39) (5, 18) (6, 13) (7, 6) (8, 8) (9, 3) (10, 1.1) (11, 2) (12, 1.1) (13, 2) (14, 1.1) (15, 1.1) (17, 1.1) (18, 1.1) (19, 1.1) (20, 2) (22, 1.1) (23, 1.1) (25, 3) (26, 2) (31, 1.1) (35, 1.1) (38, 1.1) (40, 1.1) (45, 3) (48, 1.1) (74, 2) (80, 1.1) (165, 1.1) (253, 1.1) (316, 1.1) (538, 1.1) (743, 1.1) (1040, 1.1) (1216, 1.1) (1456, 1.1) (2525, 1.1) (3953, 1.1) (7958, 1.1) (9224, 1.1) (15853, 1.1) (61386, 1.1) (71233, 1.1) (77286, 1.1)};
      \end{axis}
    \end{tikzpicture}
  }
  \vspace{-8mm}
  \captionof{figure}{\revisionPC{Histogram on the number of iterations consumed by the sample-and-check loop at \crefldecoding{line:decoding-loop2} of \cref{alg:decoding} to find a valid token, measured with Gemma 2 2B for HumanEval synthesis.}}
  \label{fig:correction-histogram}
\end{wrapfigure}
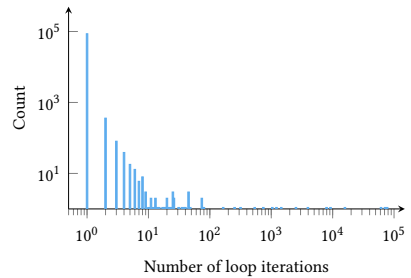

\paragraph{Number of Sample-and-Check Loop Iterations}
To provide an in-depth analysis of the overhead of our type constraining method, we measure the number of iterations spent by the sample-and-check loop to find an admissible token.
The results are provided in \cref{fig:correction-histogram}.
We observe that the number of loop iterations follows a long-tail distribution.
For $99.4\%$ of cases, only one loop iteration is needed.
This number is even higher for stronger models, with Gemma 2 9B and 27B requiring one iteration in $99.6\%$ and $99.9\%$ of cases, respectively.
This means that, in most instances, LLMs can generate a valid token on the first attempt, which is then verified by the completion engine.
In cases where more than one iteration is needed, the completion engine intervenes to guide the selection of valid tokens.
These interventions help resolve errors in many instances in our benchmarks, providing significant benefit, as discussed in \cref{sec:experiments-results}.

\revisionOC{
Prior work~\citep{ugare2024syncode,dominos-DBLP:journals/corr/abs-2403-06988,synchromesh} implemented constrained decoding differently than \cref{alg:decoding}.
Instead of running the sample-and-check loop, they execute the completion engine for all tokens in the LLM's vocabulary, mask out all invalid tokens, and sample once from the remaining valid tokens based on their normalized likelihoods.
This implementation is less efficient than ours, especially when calling the completion engine is costly.
Based on the long-tail distribution depicted in \cref{fig:correction-histogram}, our implementation requires only a single invocation to the completion engine for most cases, whereas their implementation calls it for every token in the LLM's vocabulary for all cases.
In our experiments, their implementation results in timeouts for all benchmark instances, with a timeout of 5 minutes per instance.
This justifies our choice of using the sample-and-check loop.
}

%% file: sections/casestudy.tex
\begin{figure}
  \lstset{
  basicstyle=\small,
  breaklines=true,
  language=JavaScript,
  numbers=none,
  escapechar=\%,
  basicstyle=\linespread{0.7}\ttfamily\footnotesize,
  framexleftmargin=3mm, 
  framexrightmargin=10mm,
  aboveskip=10mm,
  belowskip=10mm,
  }
  \def\arraystretch{1.3}
  \small
  \centering
  \setlength{\tabcolsep}{8pt}
    \begin{tabular}{@{}l|l|l@{}}
        &\multicolumn{1}{c|}{\small Vanilla }&\multicolumn{1}{c}{\small Types}

        \\\hline
        {(a)}&
    \begin{lstlisting}
// find all words with consonants
function select_words(s: string): string[] {
  for (const word of s.%\wrongtok{split()}%
    \end{lstlisting} &
    \begin{lstlisting}
// find all words with consonants
function select_words(s: string): string[] {
  for (const word of s.%\righttok{split(\textquotesingle\ \textquotesingle)}%
    \end{lstlisting}
        \hspace{1pt}
        \\[4mm]\hline
        {(b)}&
    \begin{lstlisting}
function largest_divisor(n: number): number {
  for(let i = n - 1; i > 0; i--) {
      if(n %\%% i === 0) return i;
  }
  %\wrongtok{// No return}%
}
    \end{lstlisting} &
    \begin{lstlisting}
function largest_divisor(n: number): number {
  for(let i = n - 1; i > 0; i--) {
      if(n %\%% i === 0) return i;
  }
  %\righttok{return 1;}%
}
    \end{lstlisting}
        \hspace{1pt}
        \\[8mm]\hline
        {(c)}&
    \begin{lstlisting}
// insert a delimiter between every element
  return numbs.reduce(
    (%\wrongtok{acc}%, curr, index) => {
        acc.push(curr);
        return acc;
  }, []);
    \end{lstlisting} &
    \begin{lstlisting}
// insert a delimiter between every element
  return numbs.reduce(
    (%\righttok{acc: number[]}%, num, index) => {
        acc.push(curr);
        return acc;
  }, []);
    \end{lstlisting}
        \hspace{1pt}
        \\[5mm]
      
  \end{tabular}

  \vspace{-2mm}
    \caption{\revisionOC{Three examples illustrating the effect of type-constrained sampling. 
    Left are unconstrained generations with problematic tokens highlighted in \wrongtok{red}, and right are type-constrained results with corrected tokens highlighted in \righttok{green}, adapted for clarity. 
    In (a), Gemma 2 2B attempts to call \ttt{split}, missing required arguments. 
          In (b), DeepSeek Coder 33B attempts to complete a function without a guaranteed return. The issue is resolved by forcing generation of another statement after the main loop.
          In (c), Gemma 2 9B calls \ttt{reduce} with an anonymous function without type annotation. This leads to an incorrect type inference for the first parameter. The issue is solved by guiding the model to add type annotation.}}
  \label{fig:case-study}
\end{figure}
\subsection{Case Study}
\label{sec:case-study}

\definecolor{mycrefcolor}{HTML}{6D2382} 

\newcommand{\colorcref}[1]{\textcolor{mycrefcolor}{#1}}

For a qualitative evaluation, we manually inspect instances where unconstrained decoding fails and our type constraining approach successfully corrects errors.
We find that our technique effectively amends various types of compilation errors.
We showcase three such examples in \cref{fig:case-study}.

\paragraph{Wrong Parameter Count}
In \cref{fig:case-study}\colorcref{a}, the task is to find all words in a string that contain a number of consonants. 
When Gemma 2 2B attempts to solve this problem, translating from the Python version to TypeScript, it calls \ttt{split} without arguments. This is allowed in Python, but not in TypeScript, where a single $\ttt{string}$ argument is required. Type constraining correctly detects this and guides the generation to sample a space character as parameter.

\paragraph{Missing Return Statement}
In \cref{fig:case-study}\colorcref{b}, to complete function \ttt{largest_divisor}, the model must compute a straightforward divisor loop. DeepSeek Coder 33B Instruct \citep{deepseek-coder-DBLP:journals/corr/abs-2401-14196} implements a correct loop, but does not guarantee returning a value in every execution path. When the return statement in the loop is never executed, e.g., for negative inputs, the function thus returns $\ttt{undefined}$, violating the type rules.
Our method detects this issue and forces the generation of another statement in the function body, resulting in a correct fallback return statement.

\paragraph{Incorrect Type Inference}
In \cref{fig:case-study}\colorcref{c}, the task is to insert a delimiter between every element in an array.
Gemma 2 9B solves this with the \ttt{reduce} function. This generic function accepts two arguments; first, a callback function that is called consecutively for every element in the array and accumulates a result, second, an initial value for the callback function.
The type of the accumulator of the callback is derived implicitly from the second argument, which is an empty array in the given example.
TypeScript infers special type \ttt{never[]} for the empty array, disallowing inserting \ttt{curr} of type \ttt{number} through \ttt{push}. Therefore, the program fails to compile.
This issue is a well-known limitation of the TypeScript compiler, often confusing even expert developers \citep{typescript-issue-59863,typescript-issue-59999}. Our method resolves it by enforcing adequate type annotation on the first argument of the callback function.

%% file: sections/discussion.tex
\section{Discussion}
\label{sec:discussion}

Our general type constraining approach, backed by strong experimental results, opens exciting avenues for future research, which we discuss below.

\paragraph{Implementation Effort}
Developing a completion engine for a target programming language currently requires manual efforts.
However, we expect that the involved effort to adopt our method to other languages will be reduced significantly, as many features transfer from our implementation for $L_B$ and TypeScript.
Moreover, we believe that, due to the huge impact on LLM's code generation, the effort will pay off.
Future programming language developers may consider generally writing their compilers as an incremental completion engine, which additionally enables automatic adoption for constrained code generation, besides conventional grammar parsing and type checking.

\revisionPC{
\paragraph{Broader Application to More Complex Tasks and Stronger LLMs}
Stronger LLMs, such as the latest OpenAI models~\citep{o1-system-card}, may make fewer typing errors on the HumanEval and MBPP datasets.
Our evaluation results in \cref{fig:main-synth-translation-res} also demonstrate that compilation errors decrease with increasing model size for the Gemma family.
However, recent findings showed that currently, even the strongest LLMs struggle with generating compilable code for more complex coding tasks, stricter typing rules, and low-resource languages (e.g., new DSLs).
\citet{compilation-benchmark} evaluated various state-of-the-art LLMs on difficult code synthesis tasks in Rust, reporting compilation error rates of $18\%$ for OpenAI o1-mini \citep{o1-system-card}, $39\%$ for DeepSeek R1 \citep{deepseek-r1} and $27\%$ for Anthropic's Claude 3.5 Sonnet \citep{anthropic-claude3-model-card}.
For OCaml and Haskell, which are sparsely represented in LLMs' training data, the error rate is even higher at $40\%-60\%$ for all models, matching a trend of worse performance on low-resource languages \citep{survey-low-resource-domain-specific-pl,enhancing-code-generation}.
\citet{lost-in-translation} compiled a large dataset of code translation and found $44.3\%$ of GPT-4-generated code to contain compilation errors. Similarly, \citet{syzygy} report around $25\%$ compilation errors for C-to-Rust translation using OpenAI o1 models.
Our type constraining approach is broadly applicable to all these scenarios and our work presents a promising proof of concept.
Future work can consider building upon our approach to address these challenges.

Constrained decoding in general requires access to the next-token probability distributions produced by LLMs.
Currently, commercially available black-box LLM APIs only return sampled tokens and do not offer complete next-token distributions.
A possible solution is to integrate our method into the backend of model providers, as was recently implemented for guaranteeing adherence to JSON Schemas \citep{anthropic-json-mode,openai-structured-outputs}.
}

\paragraph{Remaining Compiler Errors}
We observe that, even though constrained decoding guarantees a valid result upon termination, a considerable amount of compilation errors remain due to non-termination within the token or time limit. We find this to be caused by generation loops, entered when generation is amended by constraints and the LLM is unable to recover. An example is depicted in \cref{fig:non-terminating-code}, where CodeLlama 34B tries to access the invalid member \ttt{sort} on an expression of type \ttt{number}. Future work may add additional constraints to force stopping such unconstructive loops and steer the model more strictly, e.g., by limiting the complexity of generated expressions.

\input{figures/recovery}

%% file: figures/recovery.tex
\begin{figure}
    \centering
    \vspace{0cm}
    \centering
    \begin{lstlisting}[language=JavaScript,escapechar=\%]
function sort_third(l: number[], r: number[]): number[] {
    for (let i = 0; i < l.length; i++) {
        r.push(l[i]%\righttok{.toString}%().slice(0, 3).concat(l[i].toString().slice(3).split('').reverse()
        .join('')).split('').reverse().join('').toString() + l[i].toString().slice(3).split('') ...
    \end{lstlisting}
%
\vspace{-4mm}
    \caption{Compilations errors remain when the model does not terminate after a corrected token. In this example for synthesis on the HumanEval task \#33, CodeLlama 34B is steered away from accessing non-existing member \wrongtok{\ttt{.sort}} and instead accesses \righttok{\ttt{.toString}}.}
    \label{fig:non-terminating-code}
\end{figure}

%% file: sections/related.tex
\section{Related Work}
\label{sec:related-work}

\paragraph{Code Language Models}
Recently, LLMs have gained traction for diverse coding tasks such as code synthesis, repair, or translation \citep{jiang2024surveylargelanguagemodels}. These models are typically trained on datasets containing billions to trillions of tokens and have billions of parameters, with both factors contributing to improved performance in code-related benchmarks \citep{code-llama-DBLP:journals/corr/abs-2308-12950,gemma-2024,deepseek-coder-DBLP:journals/corr/abs-2401-14196,mundler2024code}. Meanwhile, LLMs are well known to frequently make mistakes \citep{rawte2023survey-hallu,huang2023survey-hallu}, and, as we show in this work, even state-of-the-art open-weight models with over 30 billion parameters frequently make errors in code generation.

\paragraph{Improving Language Model Accuracy}
Apart from constrained decoding, three primary approaches have been proposed to enhance the accuracy of language models on code tasks: fine-tuning, retrieval augmentation (RAG), and compiler or execution feedback. Fine-tuning adapts the model weights based on specifically collected training data. This process is highly resource intensive \citep{tsai2024codelessalignmore,weyssow2024exploringparameterefficientfinetuningtechniques}. 
\revisionPC{RAG provides the model with additional context based on a database or related code snippets \citep{finetuning-rag-bassamzadeh2024comparativestudydslcode,synchromesh}.}
Compiler and execution feedback is only available after completing the model generation and requires resampling \citep{cotran-Jana-2024,deligiannis2023fixingrustcompilationerrors,prog-synth-refinement-types-DBLP:conf/pldi/PolikarpovaKS16}.
\revisionPC{
However, constrained decoding is orthogonal to these methods and, as indicated by \citet{synchromesh} and our experimental results, combining constrained decoding with RAG or compiler feedback additionally improves model performance.
}


\paragraph{Constrained Decoding}
Prior work on constrained decoding failed to achieve strong results due to its limitation to syntactic language features.
Constraining to context-free languages has been explored extensively in recent work \citep{dominos-DBLP:journals/corr/abs-2403-06988,synchromesh,promptingisprogramming10.1145/3591300,willard2023efficientguidedgenerationlarge}. Simple context-sensitive syntactic features, such as the space indentation in Python and the scope markers in Go have also been implemented \citep{melcer2024constraineddecodingfillinthemiddlecode, ugare2024syncode}. As demonstrated in \cref{sec:experiments}, however, syntax errors on average account for only $6\%$ of compilation errors in recent code models. The rarity of syntax errors significantly reduces the potential of leveraging them for improvements in code correctness. Meanwhile, our type-constrained decoding more than halved compilation errors.

\paragraph{Type Systems for Code Synthesis}
Previous work that leveraged type systems for code synthesis was confined to specialized settings and unable to constrain general, complex program generation. \citet{synchromesh} proposed using known column names to guide SQL query generation. \citet{completion-DBLP:conf/pldi/GveroKKP13} employed a search on the type graph for function call completion.
\revisionPC{\citet{monitor-guided-decoding-DBLP:journals/corr/abs-2306-10763} leverage language-server-generated type annotations for object member accesses. \citet{blinn-typed-holes} use language-server-derived type information to provide additional context to the LLM, but not to enforce hard constraints.}
Additionally, type constraints have been used to direct code synthesis based on specialized search procedures \citep{prog-synth-refinement-types-DBLP:conf/pldi/PolikarpovaKS16,rust-synth-DBLP:journals/pacmpl/FialaI0PS23,partial-expression-10.1145/2254064.2254098}. However, these methods are not compatible with LLM-based code generation. This limits their ability to exploit the powerful natural language and general-purpose capabilities of LLMs. 

%% file: sections/conclusion.tex
\section{Conclusion}
\label{sec:conclusion}

In this work, we explored how type systems in programming languages can be used to guide language models during decoding. Concretely, we design and implement prefix automata to perform type constraining for a foundational simply typed language and then extend it to the popular language TypeScript. We extensively evaluate the impact of using such constraints for code synthesis, translation, and repair and observe that we more than halve compilation errors on a diverse set of models and
consistently increase functional correctness. We further explore qualitatively how the constraining positively impacts code generation. We conclude that such type constraining should be implemented for more programming languages, and has the potential to generally improve code generation in many domains.

%% file: sections/appendix.tex
\appendix

\input{sections/appendix/ce\_details}
\input{sections/appendix/experimental\_details}
\input{sections/appendix/casestudy\_details}

\clearpage

\begin{table}[p]
  \small
  \centering
  \setlength{\tabcolsep}{14pt}
  \caption{\revisionPC{Supported TypeScript features.}}
  \vspace{-3mm}
  \label{tab:feature-set}
  \begin{tabular}{@{}ll@{}}
  \toprule
  \textbf{Supported TypeScript Features} & \textbf{Examples} \\ \midrule
  Expressions, Statements, Function Declarations & ($L_B$ as introduced in \cref{sec:method}) \\
  Additional Literals: BigInt, Regex, Template Strings & \ttt{10n, /\textbackslash{}d*/, `hello \$\{user\}`}\\
  Additional Types: void, null, undefined & \ttt{void, undefined, null}\\
  Index Signature Types and Literals & \ttt{let x: \{[y: number]: string\} = {1: "hi"};}\\
  Anonymous Functions & \texttt{function (): bool \{return true\}} \\
  Lambda Functions with and without Function Bodies & \texttt{x => \{return y\}}, \texttt{x => y} \\
  Ternary and LogicOperators & \texttt{? :}, \texttt{||}, \texttt{\&\&} \\
  Arithmetic and Boolean Operations & \texttt{+}, \texttt{-}, \texttt{**}, \texttt{\&}, \texttt{!} \\
  Assigning Pre-and Postfix Operators & \texttt{++}, \texttt{---} \\
  Arrays & \ttt{[1, 2, 3]}\\
  Access and Assignment to Computed Members & \texttt{x[10] = y[i];} \\
  Constructors and "new" Calls & \ttt{let x = new Number(1);}\\
  Calls with Optional and Rest Parameters & function foo(x?: number, y...: string) \\
  Sets and Maps & \ttt{Map<string, number>()}\\
  Parameterized Constructor Calls & \texttt{new Set<string>()} \\
  Tuples & \texttt{let x : [int, string] = [1, "hello"];} \\
  Optional Chaining & \texttt{x.get("hi")?.get("world")} \\
  Spread Operator & \texttt{[...xs]} \\
  Type Assertions & \texttt{"hello" as any} \\
  For Loops & \texttt{for(int x = 0; i < 10; i++)} \\
  For Of Loops & \texttt{for(x of xs)} \\
  For Of Loops with Tuple Destructuring & \texttt{for([x, y] of xys)} \\
  Do-While and While Loops & \ttt{while (true) \{...\}}\\
  Typed and Untyped Variable Declarations & \ttt{let x: number = 1; let y = 100;}\\
  Comments, Multiline Comments & \ttt{// Comment }\\
  Returning without Expressions & \ttt{return;}\\
  Try-Catch Statements with a Fixed Exception Type & \ttt{try \{ ... \} catch (e) \{ ... \}} \\
  Throw Statements & \ttt{throw new Error("...")} \\
  Importing the \ttt{crypto} Library & \texttt{require("crypto")} \\
  Global Scope Objects & \ttt{Math, parseInt}\\
  Automatic Semicolon Insertion & \\
  \bottomrule
  \end{tabular}
\end{table}

\begin{table}[p]
  \vspace{5mm}
  \small
  \centering
  \setlength{\tabcolsep}{25pt}
  \caption{Unsupported TypeScript features.}
  \vspace{-3mm}
  \label{tab:feature-unset}
  \begin{tabular}{@{}ll@{}}
  \toprule
  \textbf{Missing Features} & \textbf{Examples} \\
  \midrule
  General Library Imports & \ttt{require("example")} \\
  Use of Functions Before Declaration & \\
  For In Loops & \texttt{for(x in y)} \\
  Type Declaration & \\
  User-Defined Classes & \\
  Declaration and Parameterized Call of General Parameterized Functions & \\
  Destructuring Assignment & \texttt{[x, y] = z} \\
  Uninitialized, Unannotated Variable Declarations & \ttt{let x;}\\
  Return Type Inference & \\
  Literal Types &  \\
  Enums & \\
  Symbols & \\
  \bottomrule
  \end{tabular}
\end{table}

\clearpage

\input{figures/prompts/synth.tex}
\input{figures/prompts/translation.tex}

\clearpage

\input{figures/prompts/repair.tex}

\clearpage

\begin{figure}
  \vspace{43mm}
  \centering
  \begin{lstlisting}[language=JavaScript,escapechar=\%,basicstyle=\linespread{1}\ttfamily\footnotesize]
```typescript
function next_smallest_palindrome(num: number): number {
const numStr = num.toString();
const reversedNumStr = numStr.split('').reverse().join('');
let min = Math.min(Number(numStr), Number(reversedNumStr));

if (min === Number(numStr)) {
  return Number(numStr);
} else {
  return parseInt(min.toString() + 1);
}
%\colorbox{myred}{\hz{\}}}%

const findNextSmallestPalindrome = next_smallest_palindrome(12321);
console.log(findNextSmallestPalindrome); // Output: 12322
%\colorbox{myblue}{\hz{\textasciigrave\textasciigrave\textasciigrave}}%

**Explanation**

1. **Input:**  The function takes a number as input (`num`) in the given criteria.
...
  \end{lstlisting}
  \vspace{-4mm}
  \caption{The code generated by Gemma 2 2B for MBPP \#0. The cutoff algorithm first extracts the fenced code, last line marked in \colorbox{myblue}{\hz{blue}}. To avoid the usage of the \ttt{console} object from the browser extension of TypeScript in the generated examples, we further cut off after the last code block, marked in \colorbox{myred}{\hz{red}}.}
  \label{fig:code-extract-example}
\end{figure}

\clearpage

\input{figures/casestudy-details.tex}

%% file: sections/appendix/ce_details.tex
\section{Detailed Prefix Automaton Definitions}
\label{app:ce-details}

In this section, we provide more detailed definitions and analysis of the various automata for \subts.

\subsection{Base Automata}
\label{app:ce-basics}
We now provide detailed definitions for the base prefix automata introduced at the end of \cref{sec:ce-def}: union, concatenation, Kleene-Star, and terminal.

\paragraph{Union}
For the union $\sautomaton{}_X \cup \sautomaton{}_Y$, we define the resulting sets of initial states and accepting states as $\initial \coloneq \initial_X \cup \initial_Y$ and $\final \coloneq \final_X \cup \final_Y$, respectively.
The transition function is defined as follows:
\[
    \trans(\sstate, \sym) \coloneq 
\begin{cases}
    \trans_X(\sstate, \sym) & \text{if } \sstate \in \states{}_X\\
    \trans_Y(\sstate, \sym) & \text{if } \sstate \in \states{}_Y.
\end{cases}
\]
To show that the language parsed by this automaton is indeed the union $L(\sautomaton_X \cup \sautomaton_Y) = L(\sautomaton_X) \cup L(\sautomaton_Y)$, we employ a short helper lemma, which can be shown inductively.
\begin{lemma}
The set of the reachable states from a set of states $\mstate$ is equal to the union of reachable states from each state in $\mstate$, i.e., $\reachf(\mstate, \istr) = \bigcup_{\sstate \in \mstate}{\reachf(\sstate, \istr)}$.
\end{lemma}

Since the states are distinct and we merely combine the transition functions of both automata, using the lemma, we can quickly see that the language parsed is indeed the union.
Moreover, if both $\sautomaton_X$ and $\sautomaton_Y$ are prefix automata, this also holds for $\sautomaton_X \cup \sautomaton_Y$.

\paragraph{Concatenation}
For the concatenation automaton $A_{XY}$, we define $\initial \coloneq \initial_X$, $\final \coloneq \final_Y$, and the transition function as follows:
\[
    \trans_{XY}(\sstate, \sym) \coloneq 
\begin{cases}
    \trans_X(\sstate, \sym) & \text{if } \sstate \in \states{}_X \backslash \final_X \\
    \trans_X(\sstate, \sym) \cup \trans_Y(\initial_Y, \sym) & \text{if } \sstate \in \final_X \\
    \trans_Y(\sstate, \sym) & \text{if } \sstate \in \states{}_Y. \\
\end{cases}
\]
Informally, concatenation preserves the parsing behavior of both $\sauto_X$ and $\sauto_Y$ in their respective states.
When $\sauto_{XY}$ reaches an accepting state of $\sauto_X$ and receives another input character, it either remains in $\sauto_X$ or transitions to $\sauto_Y$, as defined in the second case of $ \trans_{XY}$.
Essentially, this maintains outgoing edges from accepting states in $\sauto_X$ while adding edges from these accepting states to initial states of $\sauto_Y$.

It follows from a similar argument that $L(A_{XY}) = L(\sauto_X) \strcat L(\sauto_Y)$, where $L(\sauto_X) \strcat L(\sauto_Y)$ is defined as $\{\istr_X \strcat \istr_Y \mid{} \istr_X \in L(\sauto_X), \istr_Y \in L(\sauto_Y)\}$.
We first show $L(A_{XY}) \subseteq L(\sauto_X) \strcat L(\sauto_Y)$.
Due to (P1), we can always split any $\istr \in L(A_{XY})$ into $\istr_X$ that extends from $\initial_X$ to $\final_X$ and $\istr_Y$ that extends from $\initial_Y$ to $\final_Y$. Then $\istr_X \in L(\sauto_X)$ and $\istr_Y \in L(\sauto_Y)$. For $L(\sauto_X) \strcat L(\sauto_Y) \subseteq L(\sauto_X \strcat \sauto_Y)$, we pick any $\istr_X \strcat \istr_Y$ from $L(\sauto_X) \strcat L(\sauto_Y)$ and parse it using $A_{XY}$. We observe that it will first traverse from $\initial_X$ to $\final_X$ consuming $\istr_X$, and then transition through $\initial_Y$ to $\final_Y$ by consuming $\istr_Y$.

Moreover $A_{XY}$ is a prefix automaton, if $\sauto_X$ and $\sauto_Y$ are prefix automata and $L(\sauto_Y) \neq \varnothing$.
Since $\sauto_X$ is a prefix automaton, we can reach $\final_X$ from any state in $\states{}_X$.
From $\final_X$ we additionally reach $\initial_Y \subseteq \states_Y$. Since $\sauto_Y$ is a prefix automaton, we can reach $\final_Y$ for any state in $\states_Y$.
This construction is a prefix automaton only if $\initial_Y \neq \varnothing$, which, due to the prefix property, is equivalent to $L(\sauto_Y) \neq \varnothing$.

\paragraph{Kleene-Star}
We define the Kleene-Star automaton $A_{\overline{X}}$ that parses indefinite repetitions of words accepted by $X$. First, we consider all initial states as final states, i.e., we ensure $\initial_X \subseteq \final_{\overline{X}}$. Then we add transitions to the transition function $\trans_X$ from the final states $\final_X$ back to the initial states $\initial_X$.
\[
    \trans_{\overline{X}}(\sstate_X, \sym) \coloneq 
\begin{cases}
    \trans_X(\sstate_X, \sym) & \text{if } \sstate \notin \final_X \\
    \trans_X(\sstate_X, \sym) \cup \trans(\initial_X, \sym) & \text{if } \sstate_X \in \final_X. \\
\end{cases}
\]
We can quickly see that $L(\sauto_{\overline{X}}) = \lbrace \overline{\istr} \mid \istr \in L(A_X) \rbrace$, with the same argument as the concatenation automaton. Additionally, because the initial states are accepting, the empty word (zero repetitions) is in $L(\sauto_{\overline{X}})$.
We similarly see that this is prefix automaton if $\sauto_X$ is a prefix automaton.
Note that here $L(\sauto_X) \neq \varnothing$ is not required.
This is because if $L(\sauto_X) \neq \varnothing$, then $\sauto_{\overline{X}} = \sauto_X = A_{\varnothing}$, which is still a prefix automaton.

\paragraph{Terminals}
The terminal automaton $A_\ttt{S}$ parses exactly the terminal $\ttt{S}$. 
They accept the usual alphabet $\Sigma$ and feature the states $\states \coloneq \{\sstate_\ttt{s} \mid \ttt{s} \text{ is a suffix of } \ttt{S}\}$, $\final \coloneq \{\sstate_\varepsilon\}$, $\initial \coloneq \{\sstate_{\ttt{S}}\}$.
The transition function $\trans$ is defined as follows:
\[
    \trans(\sstate_{\ttt{s}}, \sym) \coloneq 
\begin{cases}
    \{\sstate_{\ttt{s\textquotesingle}}\} & \text{if } \sym \strcat \ttt{s\textquotesingle} = \ttt{s}\\
    \varnothing              & \text{otherwise}.
\end{cases}
\]
Clearly $A_{\ttt{S}}$ is a prefix automaton.
We can show inductively that for any $\ttt{s}$: $\reachf(\sstate_{\ttt{s}}, \ttt{s\textquotesingle}) = \{\sstate_\varepsilon\} \iff \ttt{s} = \ttt{s\textquotesingle}$, and thus $L(A_{\ttt{S}}) = \{\ttt{S}\}$.
With a simple modification, we introduce $A^W_{\ttt{s}}$, where $W$ denotes whitespace characters.
The transition function is defined as $\trans(\sstate^W_{\ttt{s}}, \sym) \coloneq \{\sstate^W_{\ttt{s}}\}$ if $\sym \in W$; otherwise, $\trans(\sstate^W_{\sym \strcat \ttt{s}}, \token) \coloneq \{\sstate^W_\ttt{s}\}$.
This allows arbitrary whitespaces before parsing $\ttt{s}$.
This is how we implement syntactic indifference to whitespace between terminals.

%

\subsection{Expressions}
\label{app:ce-expression-details}
Expressions are parsed using recursive automatons as introduced in \cref{sec:ce-expressions}.
In this part of the Appendix, we describe in more detail how information is passed between states.

\paragraph{Notation}


In the following, we will implicitly assume that $\trans(\sstate, \sym) = \varnothing$ if not explicitly defined otherwise, making notation more concise.
For any state, we access the following information through dot notation or the special notation on the state, which we assume is passed to subsequent states through the transition function (unless otherwise stated).
This information is alternatively passed through to entire automata in composite automata, e.g., in $A_{XY}$ from $A_X$ to $A_Y$.
\begin{itemize}[leftmargin=7mm]
  \item $\sstate \in \final_X$: Whether state $\sstate$ is an accepting state of the automaton $\sauto_X$.
  \item $\sstate.\te$: The type environment based on state $\sstate$ currently being parsed.
  \item $\sstate.\textsc{lhs}$: The left-hand side expression of an extending expression represented by state $\sstate$, i.e., when extending $X$ with $Y$ and currently parsing $\sstate_Y$, then $\sstate_Y.\textsc{lhs} = X$.
  \item $\sstate.\textsc{typ}$: The described type of the last coherent expression that this state belongs to. This is only defined for accepting states. Generally, we ensure that when some expression $e$ was parsed, the corresponding state $\sstate_e$ has attribute $\sstate_e.\textsc{typ}$ such that $\sstate_e.\te |- e : \sstate_e.\textsc{typ}$.
  \item $\sstate \downarrow T$: Type $T$ to which state $\sstate$ is constrained.
\end{itemize}
When accessing the properties of $A$, we access the property of the current state of the automaton $\sstate \in \states$, e.g., $A.\textsc{lhs} = q.\textsc{lhs}$. For parsed automata, the current state is the final, accepting state.
The \textsc{typ} attribute expresses the type of the expression parsed so far. In expression states \sstate, we leverage the \textsc{lhs} to accurately determine $\sstate.\textsc{typ}$.

\vspace{2mm}
{\small
  \setlength{\tabcolsep}{14pt}
\begin{tabular}{@{}lr@{}}
{$\!\begin{aligned}
 \sstate_{\textsc{str}}.\textsc{typ} &\coloneq \texttt{string} \\
 \sstate_{\textsc{num}}.\textsc{typ} &\coloneq \texttt{number} \\
 \sstate_{\textsc{bool}}.\textsc{typ} &\coloneq \texttt{boolean} \\
 \sstate_{x}.\textsc{typ} &\coloneq  T \text{ where } \sstate_{x}.\te |- x : T \\
 \sstate_{\ttt{(}\overline{p}\ttt{)} \ttt{=>} e}.\textsc{typ} &\coloneq \ttt{(}\sauto_{\overline{p}}.\textsc{typ}\ttt{)}\ \ttt{=>}\ \sauto_e.\textsc{typ} \\
\end{aligned}$}
&
{$\!\begin{aligned}
 \sstate_{(e)}.\textsc{typ} &\coloneq \sauto_e.\textsc{typ} \\
 \sstate_{\genop e}.\textsc{typ} &\coloneq R, \text{ for } \sstate_{\genop e}.\textsc{lhs}.\textsc{typ} = S, \sauto_e.\textsc{typ} = T \text{ and } S \genop T : R \\
 \sstate_{(\overline{e})}.\textsc{typ} &\coloneq T, \text{ for } \sstate_{(\overline{e})}.\textsc{lhs}.\textsc{typ} = \ttt{(}\overline{p}\ttt{)}\ \ttt{=>}\ T \\
 \sstate_{\ttt{.}n}.\textsc{typ} &\coloneq T, \text{ for } \textsc{lookup}(\sstate_{\ttt{.}n}.\textsc{lhs}.\textsc{typ}, n) = T\\ \\
\end{aligned}$}
\end{tabular}
}

\paragraph{Unrestricted Expressions}
The left-hand side of the currently parsed expression is used in the definition of automata for three extending expressions; arithmetic operators, function call, and member access.
The arithmetic operator automaton constrains its states to those with valid operators, i.e.:
\[
\sauto_{\genop e} \coloneq \bigcup_{\exists R : \sauto_{\genop e}.\textsc{lhs}.\textsc{typ} \genop{} T = R}{A_{\genop} (\strcat A_{e} \downarrow T)}.
\]
For function call, the automaton is only valid if the left-hand side is a function, and accepts only the valid signature.
\[
\sauto_{\lparen\overline{e}\rparen} \coloneq \begin{cases}
  A_{\ttt{(}} \strcat (A_{\overline{e}} \downarrow \sauto_{\overline{p}}.\textsc{typ}) \strcat A_{\ttt{)}} & \text{if } \sauto_{\lparen\overline{e}\rparen}.\textsc{lhs}.\textsc{typ} = \ttt{(}\overline{p}\ttt{)}\ \ttt{=>}\ T \\
  \sauto_{\varnothing} & \text{otherwise}. \\
\end{cases}
\]
Finally, the member access automaton is a union of the automata that parses the attributes of the left-hand side expression.
Or formally,
\[
A_{\ttt{.}n} \coloneq \bigcup_{\exists T: \lookup(A_{\ttt{.}n}.\textsc{lhs}.\textsc{typ},m) = T} A_{\ttt{.m}}.
\]

\paragraph{Type-Restricted Expressions}
The type-restricted versions of the automata are covered by the definitions presented in \cref{sec:ce-expressions}. We therefore do not separately list them here.

\subsection{Pruning the Type Search}
\label{app:heuristic-reachable}

We now present our heuristic for pruning the type search recursion from the prefix automaton for type-constrained expressions in \cref{sec:ce-expressions}, i.e., our implementation of \textsc{pruneSearch} at \creflreach{alg:reachability:filter-next} of \cref{alg:reachability}.
The heuristic is based on the complexity and novelty of candidate types to explore.

\begin{figure}[t]
  \centering
  \includegraphics[width=0.6\linewidth]{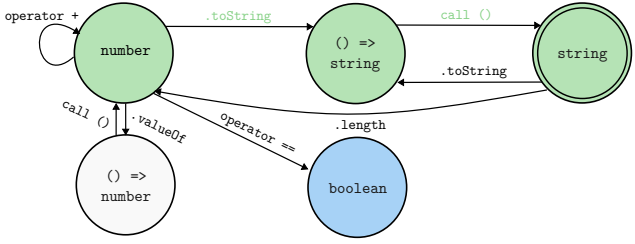}
  \vspace{-2mm}
  \caption{An example search through the graph for type reachability, starting from $T = \ttt{number}$ with the goal $\ttt{string}$, e.g., after parsing \texttt{let\;x\;:\;string;\;x\;=\;1}. States and edges along the final path are marked in \colorbox{mygreen}{\hz{green}} and explored nodes in \colorbox{myblue}{\hz{blue}}. The \ttt{()\;=>\;number} node is not explored, as complex types are avoided by our heuristic. 
  The node \ttt{()\;=>\;string} is explored as it enables reaching new type \ttt{string}.}
  \label{fig:search-graph}
\end{figure}

Based on the assumptions about the \lookup{} function and the extension expressions in \cref{sec:language}, we observe a restriction in the reachable types by extensions: from any given type, we reach itself, result types of arithmetic operators via \textsc{op}, return types through \textsc{call}, and member types through \textsc{member}. A higher-order type $\ttt{()\;=>\;} T$ does not allow access to types not reachable from $T$. Consequently, we avoid exploring such higher-order types unless the target type is of higher order, or the higher-order type offers novel, yet unexplored types. For instance, in \cref{fig:search-graph}, the type \ttt{()\;=>\;number} is not explored, because it is more complex than both the initial and goal types, \ttt{number} and \ttt{string}, and does not contain any unexplored type. Meanwhile, \ttt{()\;=>\;string} is explored, as it contains a novel \ttt{string} type.

To formalize this understanding, we introduce the concepts about the \emph{depth} and \emph{root types} of a given type, denoted as \depth{} and \roottypes{}, respectively.
\depth{} measures the complexity of a type, specifically the order of a function, while \roottypes{} returns all types of minimal depth (e.g., \ttt{string}, \ttt{number}, and \ttt{boolean}) that constitute a higher-order type.
They are defined as follows:

\begin{minipage}{\textwidth}
  \centering
  \begin{minipage}{.5\textwidth}
    \centering
    \begin{equation*}
      \depth(T) \coloneq
      \begin{cases}
      \depth(S) + 1 & \text{if } T = \ttt{(}\overline{p}\ttt{)}\,\ttt{=>}\,S\text{,} \\
      0 & \text{otherwise.}
      \end{cases}
    \end{equation*}
  \end{minipage}%
  \begin{minipage}{.5\textwidth}
    \centering
    \begin{equation*}
      \roottypes(T) \coloneq 
      \begin{cases}
      \roottypes(S) & \text{if } T = \ttt{(}\overline{p}\ttt{)}\,\ttt{=>}\,S\text{,} \\
      \{T\} & \text{otherwise.}
      \end{cases}
    \end{equation*}
  \end{minipage}
  \vspace{2mm}
\end{minipage}

We leverage \depth{} and \roottypes{} to implement $\textsc{pruneSearch}(T, G, S)$ for a current type $T$, a goal type $G$, and a type $S$ after an extension is applied on $T$. In general, if $G$ is not directly accessible from $T$, it will also not be accessible from expressions with the same root types but greater depth, such as $() => T$.
When $G$ is of higher order, exploring up to the depth of $G$ can be required, such as when $G = \ttt{()\;=>\;(()\;=>\;number)}$.
Based on these two ideas, we stop exploring $S$ when $\depth(S) > \max(\depth(G), \depth(T))$.

Further, if a higher-depth function returns an unexplored type, we need to explore it. Sticking to the example in \cref{fig:search-graph}, type \ttt{number} has the member \texttt{toString} of type \ttt{()\;=>\;string}. The type \ttt{string} can only be reached by exploring the member access at depth $1$.
On the contrary, we do not explore a higher-depth function if it does not introduce novel types other than those explored.
To achieve this, we adapt \cref{alg:reachability} to additionally define a set of root types $R$, which is initialized to an empty set and is updated by $R \coloneq R \cup \roottypes(T)$.
We do not explore $S$ if $ \roottypes(S) \subseteq R$.

Taking the conjunction of the aforementioned two aspects, our pruning heuristic is implemented as $\textsc{pruneSearch}(T, G, S) \coloneq \depth(S) > \max(\depth(T), \depth(S)) \land \roottypes(S) \subseteq R$.
The restrictions based on depth and root types are based on the results of the rigorously analyzed search over succinct types by \citet{completion-DBLP:conf/pldi/GveroKKP13}. This provides a robust heuristic for exploring as many relevant inhabitable types as possible. However, due to the additional complexity introduced by the lookup function, we can not guarantee completeness and instead refer to the strong empirical results in our evaluation in \cref{sec:experiments} as evidence of the search's high coverage.

\subsection{\revisionPC{Implementation of \der{}}}
\label{app:ce-recursion}
Recall that in \cref{fig:subset-derivable-def}, \der{} for function expressions are defined as: $\derivable(\sstate_{\ttt{(}\overline{p}\ttt{)}\;\texttt{=>}\;e}) \coloneq \{ \ttt{(}\overline{p}\ttt{)}\ \ttt{=>}\ T \mid{} \reachable(\der(\sstate_e), T)\}$.
This involves constructing a type reachability graph and collecting all types $T$ reachable from $\der(\sstate_e)$.
However, this process is intractable because $T$ can be of arbitrarily high-order, as such there are infinitely many $T$ to explore.
A similar issue exists for grouped expressions, as their \der{} function is also defined to enumerate reachable types.
We introduce two optimization heuristics to address this problem.

We first observe that \der{} is always called within the context of an invocation of $\reachable$ with target type $G$, e.g., $\reachable(\der(\sstate_{\ttt{(}\overline{p}\ttt{)}\;\texttt{=>}\;e}), G)$ for function expressions.
To compute $\derivable(\sstate_{\ttt{(}\overline{p}\ttt{)}\;\texttt{=>}\;e})$, we enumerate all types present on the type graph represented by $\reachable(\derivable(\sstate_e), G)$, which is finite due to application of the pruning heuristics in \cref{app:heuristic-reachable}.
In other words, we bound the maximum complexity of considered types $T$ using the pruning heuristic for reachability of target type $G$.
This leads to a sound but potentially incomplete version of \der{}.
However, since the final goal is to reach $G$, this heuristic provides a practically useful set of all relevant derivable types.

Second, we observe that the resulting two-tiered call $\reachable(\der(\sstate_{\ttt{(}\overline{p}\ttt{)}\;\texttt{=>}\;e}), G)$ can be integrated into a single call to further reduce the amount of explored types.
Concretely, when discovering some type $M$ in $\reachable(\derivable(\sstate_e), G)$, as per the previous heuristic,  we allow transitioning directly to $\reachable(\ttt{(}\overline{p}\ttt{)}\;\ttt{=>}\;M, G)$ to allow a depth-prioritizing exploration of the search graph. This allows us to efficiently discover a path to $G$ if it exists.

\subsection{Statements}
\label{app:ce-statements-details}

We define the remaining automata to capture the complete language from \cref{sec:language}.
To correctly handle function return types, we pass on related information when entering function bodies:
\begin{itemize}[leftmargin=7mm]
  \item $\sstate.R$: The expected return type of the current state $\sstate$.
  \item $\sstate.\textsc{returned}$: Whether the currently parsed program block has returned in all branches.
  \item $\sstate.\textsc{mustReturn}$: Whether the currently parsed program block must return (i.e., If-Then-Else branches do not need to contain return statements even if a return type is expected of the surrounding code block).
\end{itemize}

The single statement automaton is another recursive definition, since some statements, e.g., If-Then-Else, can themselves contain statements.
The statement automaton is defined recursively as $A_s \coloneq A_{\textsc{decl}} \cup A_{\textsc{expr}} \cup A_{\textsc{ret}} \cup A_{\textsc{block}} \cup A_{\textsc{fun}} \cup A_{\textsc{ite}}$.
The expression statement automaton and block automaton are simply defined as $A_{\textsc{expr}} \coloneq A_{e;}$ and $A_{\textsc{block}} \coloneq A_{\{\overline{s}\}}$.
The declaration automaton $A_{\textsc{decl}} \coloneq A_{\ttt{let}\;x \ttt{:} T \ttt{;}}$ captures variable names $x$ using an automaton for non-existing identifiers, which works the same way as $A_x$ except that it rejects terminals that match an existing variable. This automaton is a prefix automaton as well, since indefinite additional characters can be added to the variable name and there are only finitely many defined variables.
The If-Then-Else automaton is defined using standard concatenation:  $A_{\textsc{ite}} \coloneq A_{\ttt{if(}e\ttt{)}\;s\;\ttt{else}\;s}$.
The statements automaton $A_{\overline{s}}$, based on the Kleene-Star automaton definition and the single statement automaton.
Return statements are only non-empty when the expected return type is set, i.e. when parsing inside a function:
\[
  A_{\textsc{ret}} \coloneq 
\begin{cases}
 A_{\ttt{return }} \strcat A_e \downarrow T & \text{ if } A_{\textsc{ret}}.R = T\\
 A_{\varnothing} & \text{otherwise}.\\
\end{cases}
\]

For functions, the automaton is based on the standard concatenation  $A_{\textsc{fun}} \coloneq A_{\ttt{function}\;x\;\ttt{(}\overline{p}\ttt{)}\;:\;T\;\{\overline{s}\}}$. However, the transition function updates the states of the statement automata inside the function:
\begin{itemize}[leftmargin=7mm]
  \item $\sstate.R \coloneq T$, i.e., the return type of these statements is set to the return type of the function. This value is propagated recursively to all sub-automata.
  \item $\sstate.\textsc{mustReturn} \coloneq \ttt{true}$, for the outermost statement block automaton. It is set to false for deeper nested statement blocks and as soon as a parsed statement $X$ has $\sstate_X.\textsc{returned}$ set to true - i.e. one of the main body statements returned in every branch.
  \item $\sstate.\textsc{returned} \coloneq \ttt{false}$, per default in every statement, except a) in return automata, b) inside a multi-statement automaton where the previous statement has $\textsc{returned} = \ttt{true}$ and c) in ITE-automata where both branching statements have $\textsc{returned} = \ttt{true}$.
\end{itemize}

As long as a state $\sstate$ in a multi-statement automaton has $X.\textsc{returned} = \ttt{false}$ and $\sstate.\textsc{mustReturn} = \ttt{true}$, it can not accept but instead forces the generation of another statement. Since we can always express the requested type through literals and can always generate a return statement to fulfill this requirement, the prefix automaton property is not violated.

%% file: sections/appendix/experimental_details.tex
\section{Details about Experimental Evaluation}
\label{app:experiments-details}
In this section, we detail how executable code is extracted from the model responses and a slight modification to the decoding algorithm used, that increases throughput heuristically.

\revisionPC{
\paragraph{Implementation Details}

We have two main external dependencies. To implement the regular-expression-based literal automata, we leverage the \ttt{regex} library,\footnote{\url{https://pypi.org/project/regex/}} as it allows checking if the current string can be completed to match a regular expression. To implement LLM inference, we leverage the \ttt{transformers} library.\footnote{\url{https://huggingface.co/docs/transformers}}
We provide an exhaustive list of supported and unsupported features of the TypeScript language in our final implementation in \cref{tab:feature-set,tab:feature-unset}, respectively.
}

\paragraph{Hyperparameters}
We run the models on A100 NVidia GPUs with \SI{80}{GB} of VRAM and CUDA version 12.4.
We set the sampling temperature to 1. We set seeds to 0 to 4 on the four HumanEval runs and 0 on the one MBPP run, respectively.
We limit the completions to 1000 tokens and time out after 300 seconds.
We compute syntactic correctness using the Oxidation toolchain \citep{oxc} as the official TypeScript compiler does not clearly distinguish between syntactic and semantic errors.

\paragraph{Excluded MBPP Instances}

We discovered that a number of TypeScript translations in the MultiPL-E dataset \citep{multiple-DBLP:journals/tse/CassanoGNNPPYZAFGGJ23} contained invalidly generated nested tuples. 
\revisionOC{
After reporting them to the developers, they have been resolved in the latest version of MBPP and we include them in our evaluation.
}
Still, we find that the TypeScript translation of a number of MBPP instances contains too broad type annotation, annotating elements as \ttt{any} or array of \ttt{any}. We therefore exclude the following 6 instances from the evaluation:

\vspace{2mm}
\begin{minipage}{\textwidth}
\begin{minipage}{.4\textwidth}
\begin{itemize}
\item \texttt{mbpp_405_check_tuplex}
\item \texttt{mbpp_563_extract_values}
\item \texttt{mbpp_580_extract_even}
\end{itemize}
\end{minipage}
\begin{minipage}{.4\textwidth}
\begin{itemize}
\item \texttt{mbpp_612_merge}
\item \texttt{mbpp_725_extract_quotation}
\item \texttt{mbpp_791_remove_nested}
\end{itemize}
\end{minipage}
\end{minipage}

\paragraph{Complete Prompts}

We provide the complete LLM prompts for our evaluated tasks (synthesis, translation, and repair) in \cref{figure:full-prompt-synth,figure:full-prompt-translation,figure:full-prompt-repair}.
The prompts are templates, instantiated with instructions specific to each task and problem instance.
If system prompts are not available for a given LLM, we prepend the system prompt to the first user prompt.
The model completion starts from a pre-filled function signature, enabling unified unit testing.
For the repair prompt, we add the non-compilable model output as assistant output and use a second turn to pass back compiler outputs.
Compiler errors contain line numbers for localization, so we annotate the output with line numbers.
We find that Qwen2.5 32B tends to always generate test cases, which leads to errors during compilation.
We therefore append the sentence \ttt{Do not include test cases in the code.} to its prompt.

\paragraph{Extracting Output Code}
Given our prompts, LLMs are expected to output the resulting programs.
However, they often produce additional outputs, such as generated test cases and explanations.
Now we describe our heuristics for extracting the generated code.
We first extract the corresponding TypeScript code block (i.e., \ttt{\textasciigrave\textasciigrave\textasciigrave{}typescript\dots\textasciigrave\textasciigrave\textasciigrave}), or do not cut off if the block is not closed. Inside the code block, we cut off after the closing curly brace of the last balanced pair of curly braces, if it is followed by a newline or semicolon. This determines the last statement block generated, and avoids cutting off, e.g., inside a template literal. Again, if no such case is found, we do not prune the output. We demonstrate the operation of our cutoff heuristics in \cref{fig:code-extract-example}.

%% file: sections/appendix/casestudy_details.tex
\section{Case Study Full Outputs}
\label{app:case-study-details}

In \cref{sec:case-study}, we present the shortened versions of three qualitative examples showcasing the effectiveness of our approach.
In \cref{fig:full-case-study-a,fig:full-case-study-b,fig:full-case-study-c}, we provide the full code outputs of these examples, with detailed descriptions in the respective captions.

%% file: figures/prompts/synth.tex
\begin{figure}
    \begin{minipage}{\textwidth}
      \begin{minipage}{\textwidth}
        \centering
        \fbox{\footnotesize
        \begin{minipage}{0.95\textwidth}
          \ttfamily
          \begin{tabular}{@{}l@{}}
            System:\\
            \ \ You are an expert in TypeScript programming. Solve the given problem by writing solution code\\
            \ \ in TypeScript. When answering, insert the solution code in a \textasciigrave\textasciigrave\textasciigrave{}typescript...\textasciigrave\textasciigrave\textasciigrave{} block.\\
            \ \ \colorbox{myred}{\hz{Do not include test cases in the code.}}\\
            \\
            User:\\
            \ \ \colorbox{mygreen}{\hz{Check if in given array of numbers, are any two numbers closer to each other than}}\\
            \ \ \colorbox{mygreen}{\hz{given threshold.}}\\
            \ \ \colorbox{mygreen}{\hz{\textgreater{}\textgreater{}\textgreater{} has_close_elements([1.0, 2.0, 3.0], 0.5)}}\\
            \ \ \colorbox{mygreen}{\hz{false}}\\
            \ \ \colorbox{mygreen}{\hz{\textgreater{}\textgreater{}\textgreater{} has_close_elements([1.0, 2.8, 3.0, 4.0, 5.0, 2.0], 0.3)}}\\
            \ \ \colorbox{mygreen}{\hz{true function}}\\
            \\
            Assistant:\\
            \ \ \textasciigrave\textasciigrave\textasciigrave{}typescript\\
            \ \ \colorbox{mygreen}{\hz{function has_close_elements(numbers: number[], threshold: number): boolean \{ }}\\
          \end{tabular}
        \end{minipage}}
        \vspace{-2.5mm}
        \caption{The full prompt for the synthesis task. Text in \colorbox{mygreen}{\hz{green}} is based on the problem instance, in this case HumanEval \#0. The \colorbox{myred}{\hz{red}} sentence is only added for Qwen2.5 32B due to its tendency to generate non-compiling test cases.}
        \label{figure:full-prompt-synth}
      \end{minipage}
    \end{minipage}
\end{figure}

%% file: figures/prompts/translation.tex
\begin{figure}
    \begin{minipage}{\textwidth}
      \begin{minipage}{\textwidth}
        \centering
        \fbox{\footnotesize
        \begin{minipage}{0.95\textwidth}
          \ttfamily
          \begin{tabular}{@{}l@{}}
            System:\\
            \ \ You are a helpful and expert programmer in Python and TypeScript. You will be given an input\\
            \ \ program in Python and your task is to translate this program into TypeScript. You may assume\\
            \ \ that the input program is correct and that the translation should be semantically equivalent.\\
            \ \ When answering, insert the solution code in a \textasciigrave\textasciigrave\textasciigrave{}typescript...\textasciigrave\textasciigrave\textasciigrave{} block.\\
            \ \ \colorbox{myred}{\hz{Do not include test cases in the code.}}\\
            \\
            User:\\
            \ \ The following is the source program in Python:\\
            \ \ \textasciigrave\textasciigrave\textasciigrave{}python\\
            \ \ \colorbox{myblue}{\hz{from typing import List}}\\
            \ \ \colorbox{myblue}{\hz{}}\\
            \ \ \colorbox{myblue}{\hz{def\ has_close_elements(numbers:\ List[float],\ threshold:\ float)\ ->\ bool:}}\\
            \ \ \colorbox{myblue}{\hz{\ \ \ \ for\ idx,\ elem\ in\ enumerate(numbers):}}\\
            \ \ \colorbox{myblue}{\hz{\ \ \ \ \ \ \ \ for\ idx2,\ elem2\ in\ enumerate(numbers):}}\\
            \ \ \colorbox{myblue}{\hz{\ \ \ \ \ \ \ \ \ \ \ \ if\ idx\ !=\ idx2:}}\\
            \ \ \colorbox{myblue}{\hz{\ \ \ \ \ \ \ \ \ \ \ \ \ \ \ \ distance\ =\ abs(elem\ -\ elem2)}}\\
            \ \ \colorbox{myblue}{\hz{\ \ \ \ \ \ \ \ \ \ \ \ \ \ \ \ if\ distance\ <\ threshold:}}\\
            \ \ \colorbox{myblue}{\hz{\ \ \ \ \ \ \ \ \ \ \ \ \ \ \ \ \ \ \ \ return\ True}}\\
            \ \ \colorbox{myblue}{\hz{\ \ \ \ }}\\
            \ \ \colorbox{myblue}{\hz{\ \ \ \ return\ False}}\\
            \ \ \textasciigrave\textasciigrave\textasciigrave{}\\
            \ \ Please translate the source program to TypeScript.\\
            \\
            Assistant:\\
            \ \ \textasciigrave\textasciigrave\textasciigrave{}typescript\\
            \ \ \colorbox{mygreen}{\hz{function has_close_elements(numbers: number[], threshold: number): boolean \{ }}\\
          \end{tabular}
        \end{minipage}}
        \vspace{-2.5mm}
        \caption{The full prompt for the translation task. Text in \colorbox{mygreen}{\hz{green}} is based on the problem instance, \colorbox{myblue}{\hz{blue}} is the original Python canonical solution, in this case HumanEval \#0. The \colorbox{myred}{\hz{red}} sentence is only added for Qwen2.5 32B due to its tendency to generate non-compiling test cases.}
        \label{figure:full-prompt-translation}
      \end{minipage}
    \end{minipage}
\end{figure}

%% file: figures/prompts/repair.tex
\begin{figure}
  \vspace{7mm}
    \begin{minipage}{\textwidth}
      \begin{minipage}{\textwidth}
        \centering
        \fbox{\footnotesize
        \begin{minipage}{0.95\textwidth}
          \ttfamily
          \begin{tabular}{@{}l@{}}
            System:\\
            \ \ You are an expert in TypeScript programming. Solve the given problem by writing solution code\\
            \ \ in TypeScript. When answering, insert the solution code in a \textasciigrave\textasciigrave\textasciigrave{}typescript...\textasciigrave\textasciigrave\textasciigrave{} block.\\
            \\
            User:\\
            \ \ \colorbox{mygreen}{\hz{Given an array of positive integers x. return a sorted array of all}}\\
            \ \ \colorbox{mygreen}{\hz{elements that hasn't any even digit.}}\\
            \ \ \colorbox{mygreen}{\hz{Note: Returned array should be sorted in increasing order.}}\\
            \ \ \colorbox{mygreen}{\hz{For example:}}\\
            \ \ \colorbox{mygreen}{\hz{>>> unique_digits([15, 33, 1422, 1])}}\\
            \ \ \colorbox{mygreen}{\hz{[1, 15, 33]}}\\
            \ \ \colorbox{mygreen}{\hz{>>> unique_digits([152, 323, 1422, 10])}}\\
            \ \ \colorbox{mygreen}{\hz{[]}}\\
            \\
            Assistant:\\
            \ \ \textasciigrave\textasciigrave\textasciigrave\\
            \ \ \colorbox{myblue}{\hz{001:\ function\ unique_digits(x:\ number[]):\ number[]\ \{}}\\
            \ \ \colorbox{myblue}{\hz{002:\ \ \ const\ filteredArray:\ number[]\ =\ [];\ //\ initialize\ empty\ array}}\\
            \ \ \colorbox{myblue}{\hz{003:\ \ \ for\ (let\ i\ =\ 0;\ i\ <\ x.length;\ i++)\ \{}}\\
            \ \ \colorbox{myblue}{\hz{004:\ \ \ \ \ let\ strx\ =\ ""\ +\ x[i];}}\\
            \ \ \colorbox{myblue}{\hz{005:\ \ \ \ \ let\ filteredStr\ =\ strx}}\\
            \ \ \colorbox{myblue}{\hz{006:\ \ \ \ \ \ \ \ \ .toString()}}\\
            \ \ \colorbox{myblue}{\hz{007:\ \ \ \ \ \ \ \ \ .split("")}}\\
            \ \ \colorbox{myblue}{\hz{008:\ \ \ \ \ \ \ \ \ .filter((digit)\ =>\ parseInt(digit)\ \%\ 2\ !==\ 0);}}\\
            \ \ \colorbox{myblue}{\hz{009:\ \ \ \ \ if\ (filteredStr.length\ >\ 0)\ \{}}\\
            \ \ \colorbox{myblue}{\hz{010:\ \ \ \ \ \ \ filteredArray.push(...filteredStr);}}\\
            \ \ \colorbox{myblue}{\hz{011:\ \ \ \ \ \}}}\\
            \ \ \colorbox{myblue}{\hz{012:\ \ \ \}}}\\
            \ \ \colorbox{myblue}{\hz{013:\ \ \ return\ filteredArray.sort((a,\ b)\ =>\ a\ -\ b);}}\\
            \ \ \colorbox{myblue}{\hz{014:\ \}}}\\
            \ \ \textasciigrave\textasciigrave\textasciigrave{}\\
            \\
            User:\\
            \ \ This output produced an error:\\
            \ \ \colorbox{myred}{\hz{../../../../../tmp/tmpaewlckfv.ts(10,26): error TS2345: Argument of type 'string' is not}}\\
            \ \ \colorbox{myred}{\hz{assignable to parameter of type 'number'.}}\\
            \\
            \ \ Write the program again, and make sure to fix the error this time.\\
            \\
            Assistant:\\
            \ \ \textasciigrave\textasciigrave\textasciigrave{}typescript\\
            \ \ \colorbox{mygreen}{\hz{function unique_digits(x: number[]): number[] \{}}\\
          \end{tabular}
        \end{minipage}}
        \vspace{-2.5mm}
        \caption{The full prompt for the repair task, where we simulate a multi-turn interaction. Text in \colorbox{mygreen}{\hz{green}} is based on the problem instance. \colorbox{myblue}{\hz{blue}} is non-compilable model output from the synthesis step, in this case HumanEval \#104 and Gemma 2 2B. \colorbox{myred}{\hz{red}} is compiler feedback. Generations from the synthesis step are annotated with line numbers to aid localizing the compiler error.}
        \label{figure:full-prompt-repair}
        \vspace{4mm}
      \end{minipage}
    \end{minipage}
\end{figure}

%% file: figures/casestudy-details.tex
  \begin{figure}[hp!]
    \vspace{8mm}
    \centering
    \begin{subfigure}[t]{\textwidth}
    \begin{lstlisting}[language=JavaScript,escapechar=\%,basicstyle=\linespread{1}\ttfamily\footnotesize]
function select_words(s: string, n: number): string[] {
  const result: string[] = [];
  for (const word of s.%\wrongtok{split()}%) {
    let nConsonants = 0;
    for (let i = 0; i < word.length; i++) {
      const char = word[i].toLowerCase();
      if (!['a', 'e', 'i', 'o', 'u'].includes(char)) {
        nConsonants++;
      }
    }
    if (nConsonants === n) {
      result.push(word);
    }
  }
  return result;
}
    \end{lstlisting}
    \vspace{-2mm}
        \caption{The code generated by the unconstrained Gemma 2 2B when translating a Python solution. The model tries to call the \ttt{split} member of a \ttt{string} object without parameters. This is legal in Python but not allowed in TypeScript.}
        \vspace{4mm}
    \end{subfigure}
    \medskip
    \begin{subfigure}[t]{\textwidth}
    \begin{lstlisting}[language=JavaScript,escapechar=\%,basicstyle=\linespread{1.2}\ttfamily\footnotesize]
function select_words(s: string, n: number): string[] {
  const result: string[] = [];
  for (const word of s.%\righttok{split(\textquotesingle\ \textquotesingle)}%) {
    let nConsonants = 0;
    for (let i = 0; i < word.length; i++) {
      const letter = word[i].toLowerCase(); 
      if (!['a', 'e', 'i', 'o', 'u'].includes(letter)) {
        nConsonants++;
      }
    }
    if (nConsonants === n) {
      result.push(word);
    }
  }
  return result;
}
    \end{lstlisting}
    \vspace{-2mm}
    \caption{The code generated by Gemma 2 2B with type constraints. The constraints guide the model and help it generate the required argument to \ttt{split}, which determines the characters around which to split the string.}
    \end{subfigure}
    \caption{Comparison between the function generated by unconstrained Gemma 2 2B~\citep{gemma-2024} and the code generated with the guidance of type constraints on the problem \texttt{select_words} in HumanEval~\citep{humaneval}. This problem requires a function that selects words from a given list that contain a specified number of consonants.}
      \label{fig:full-case-study-a}
\end{figure}

\pagebreak

\begin{figure}[hp!]
    \vspace{42mm}
    \centering
    \setlength{\tabcolsep}{2pt}
    \begin{subfigure}[t]{\textwidth}
    \begin{lstlisting}[language=JavaScript,escapechar=\%,basicstyle=\linespread{1}\ttfamily\footnotesize]
function largest_divisor(n: number): number {
  for(let i = n - 1; i > 0; i--) {
      if(n %\%% i === 0) {
          return i;
      }
  }
%\wrongtok{\}}%
    \end{lstlisting}
    \vspace{-2mm}
    \caption{The code generated by the unconstrained Deepseek Coder 33B Instruct. The function loops over all numbers smaller than \ttt{n} to determine the largest divisor. Even though functionally correct for positive numbers, this function returns \ttt{undefined} when the return statement in the loop is never called (e.g., for negative inputs), violating the type system rules.}
    \label{fig:full-case-study-b1}
        \vspace{4mm}
    \end{subfigure}
    \medskip
    \begin{subfigure}[t]{\textwidth}
    \begin{lstlisting}[language=JavaScript,escapechar=\%,basicstyle=\linespread{1}\ttfamily\footnotesize]
function largest_divisor(n: number): number {
  for(let i = n - 1; i > 0; i--) {
      if(n %\%% i === 0) {
          return i;
      }
  }
  %\righttok{return 1; // Return 1 if no such divisor found}%
%\righttok{\}}%
    \end{lstlisting}
    \vspace{-2mm}
    \caption{The code generated by Deepseek Coder 33B Instruct with type constraints. The system ensures that every path returns a value matching the annotated type. This forces the LLM to generate another line after the loop, and the model correctly chooses to generate a fallback return statement.}
    \label{fig:full-case-study-b2}
    \end{subfigure}
    \caption{Comparison between the function generated by the unconstrained DeepSeek Coder 33B Instruct \citep{deepseek-coder-DBLP:journals/corr/abs-2401-14196} and the code generated with the guidance of type constraints on the problem \texttt{largest_divisor} in HumanEval~\citep{humaneval}. The problem asks to compute the largest divisor of a given number.}
    \label{fig:full-case-study-b}
\end{figure}

\pagebreak

\begin{figure}[hp!]
    \vspace{25mm}
    \centering
    \setlength{\tabcolsep}{2pt}
    \begin{subfigure}[t]{\textwidth}
    \begin{lstlisting}[language=JavaScript,escapechar=\%,basicstyle=\linespread{1}\ttfamily\footnotesize]
function intersperse(numbers: number[], delimeter: number): number[] {
  return numbers.reduce((%\wrongtok{acc}%, curr, index) => {
    acc.push(curr);
    if (index < numbers.length - 1) {
      acc.push(delimeter);
    }
    return acc;
  }, []);
}
    \end{lstlisting}
    \vspace{-2mm}
    \caption{The code generated by the unconstrained Gemma 2 9B. The code uses the \ttt{reduce} member of arrays in TypeScript. The \ttt{reduce} member is a generic function that accepts a callback as first argument. This callback has three parameters: an accumulator, a current value and an index. The types of the second and third parameter are fixed to the element type of the callee array and $\ttt{number}$. The first argument type of the callback function is automatically inferred from the type of the second argument to the \ttt{reduce} function, in this case an empty array. The type of this array is $\ttt{never}[]$, a special type denoting arrays without members. Since we can assign this type to the callee type $\ttt{number}[]$, the accumulator paramter obtains type $\ttt{never}[]$. However, calling \ttt{push} on this type results in a type violation as the pushed value of type $\ttt{number}$ is not compatible with the inferred element type $\ttt{never}$. This issue is well known in TypeScript and even human developers frequently stumble upon this issue \citep{typescript-issue-59863,typescript-issue-59999}.}
        \vspace{4mm}
    \end{subfigure}
    \medskip
    \begin{subfigure}[t]{\textwidth}
    \begin{lstlisting}[language=JavaScript,escapechar=\%,basicstyle=\linespread{1}\ttfamily\footnotesize]
function intersperse(numbers: number[], delimeter: number): number[] {
  return numbers.reduce((%\righttok{acc: number[]}%, num: number, index: number) => {
    if (index > 0) {
      acc.push(delimeter);
    }
    acc.push(num);
    return acc;
  }, []);
}
    \end{lstlisting}
    \vspace{-2mm}
    \caption{The code generated by Gemma 2 9B \citep{gemma-2024} with type constraints. We force the model to generate type annotations for the callback function in \ttt{reduce} and thus enable the model to generate the correct type annotation and avoid the issue in the unconstrained code.}
    \end{subfigure}
    \caption{Comparison between the function generated by the unconstrained Gemma 2 9B \citep{qwen2025qwen25technicalreport} and the code generated with the guidance of type constraints on the problem \texttt{intersperse} in HumanEval \citep{humaneval}. The task in this problem is to insert a delimiter number between consecutive elements of an input list.}
    \label{fig:full-case-study-c}
\end{figure}